\documentclass[sigconf]{acmart}
\AtBeginDocument{%
  }

\usepackage{listings}
\usepackage[ruled,linesnumbered,noend]{algorithm2e}
\usepackage{graphicx}
\usepackage{subcaption}
\usepackage{listings}
\usepackage{tablefootnote}
\usepackage{enumitem}
\usepackage{amsfonts}
\usepackage{xcolor}
\usepackage{multirow}

\newtheorem{definition}{Definition}

\setcopyright{acmlicensed}
\copyrightyear{2025}
\acmYear{2025}
\acmDOI{10.1145/3746027.3755043}
\acmConference[MM '25]{Proceedings of the 33rd ACM International Conference on Multimedia}{October 27--31, 2025}{Dublin, Ireland.}
\acmBooktitle{Proceedings of the 33rd ACM International Conference on Multimedia (MM '25), October 27--31, 2025, Dublin, Ireland}
\acmISBN{979-8-4007-2035-2/2025/10}

\begin{document}

\title{Querying Autonomous Vehicle Point Clouds: Enhanced by 3D Object Counting with CounterNet}

\author{Xiaoyu Zhang}
\affiliation{%
  \institution{RMIT University}
  \city{Melbourne}
  \country{Australia}}
\email{s3500791@student.rmit.edu.au}

\author{Zhifeng Bao}
\authornote{The work was done while at RMIT.}
\affiliation{
  \institution{The University of Queensland}
  \city{Brisbane}
  \country{Australia}}
\email{zhifeng.bao@uq.edu.au}

\author{Hai Dong}
\authornote{Hai Dong is the corresponding author.}
\affiliation{%
 \institution{RMIT University}
 \city{Melbourne}
 \country{Australia}}
\email{hai.dong@rmit.edu.au}

\author{Ziwei Wang}
\affiliation{%
  \institution{CSIRO}
  \city{Brisbane}
  \country{Australia}
}
\email{ziwei.wang@data61.csiro.au}

\author{Jiajun Liu}
\affiliation{%
  \institution{CSIRO}
  \city{Brisbane}
  \country{Australia}}
\email{jiajun.liu@csiro.au}

\begin{abstract}
  Autonomous vehicles generate massive volumes of point cloud data, yet only a subset is relevant for specific tasks such as collision detection, traffic analysis, or congestion monitoring. Effectively querying this data is essential to enable targeted analytics. In this work, we formalize point cloud querying by defining three core query types: RETRIEVAL, COUNT, and AGGREGATION, each aligned with distinct analytical scenarios. All these queries rely heavily on accurate object counts to produce meaningful results, making precise object counting a critical component of query execution. Prior work has focused on indexing techniques for 2D video data, assuming detection models provide accurate counting information. However, when applied to 3D point cloud data, state-of-the-art detection models often fail to generate reliable object counts, leading to substantial errors in query results. To address this limitation, we propose CounterNet, a heatmap-based network designed for accurate object counting in large-scale point cloud data. Rather than focusing on accurate object localization, CounterNet detects object presence by finding object centers to improve counting accuracy. We further enhance its performance with a feature map partitioning strategy using overlapping regions, enabling better handling of both small and large objects in complex traffic scenes. To adapt to varying frame characteristics, we introduce a per-frame dynamic model selection strategy that selects the most effective configuration for each input. Evaluations on three real-world autonomous vehicle datasets show that CounterNet improves counting accuracy by 5\% to 20\% across object categories, resulting in more reliable query outcomes across all supported query types.
\end{abstract}

\ccsdesc[500]{Computing methodologies~Computer vision problems}

\keywords{Point Cloud, Aggregation Query, Autonomous Vehicle}

\received{20 February 2007}
\received[revised]{12 March 2009}
\received[accepted]{5 June 2009}

\maketitle

\section{Introduction}
The rise of autonomous driving has significantly increased the importance of point cloud data analysis, as autonomous vehicles heavily rely on point clouds to perceive their surroundings accurately. While vast amounts of data are collected, 3D detection alone can generate approximately 5TB of point cloud data per hour~\cite{kazhamiaka2021challenges}, not all data is relevant to every training task. For instance, detecting vehicle collisions and analyzing traffic congestion differ in their focus. Collision detection often involves identifying sudden, localized changes in vehicle count within short time frames, while congestion analysis requires aggregating vehicle counts over longer durations to identify sustained traffic buildup. These tasks, therefore, depend on different object types, time scales, and aggregation strategies, reinforcing the need for flexible point cloud analytics designed for specific use cases.

\begin{figure}[t]
    \includegraphics[scale=0.32]{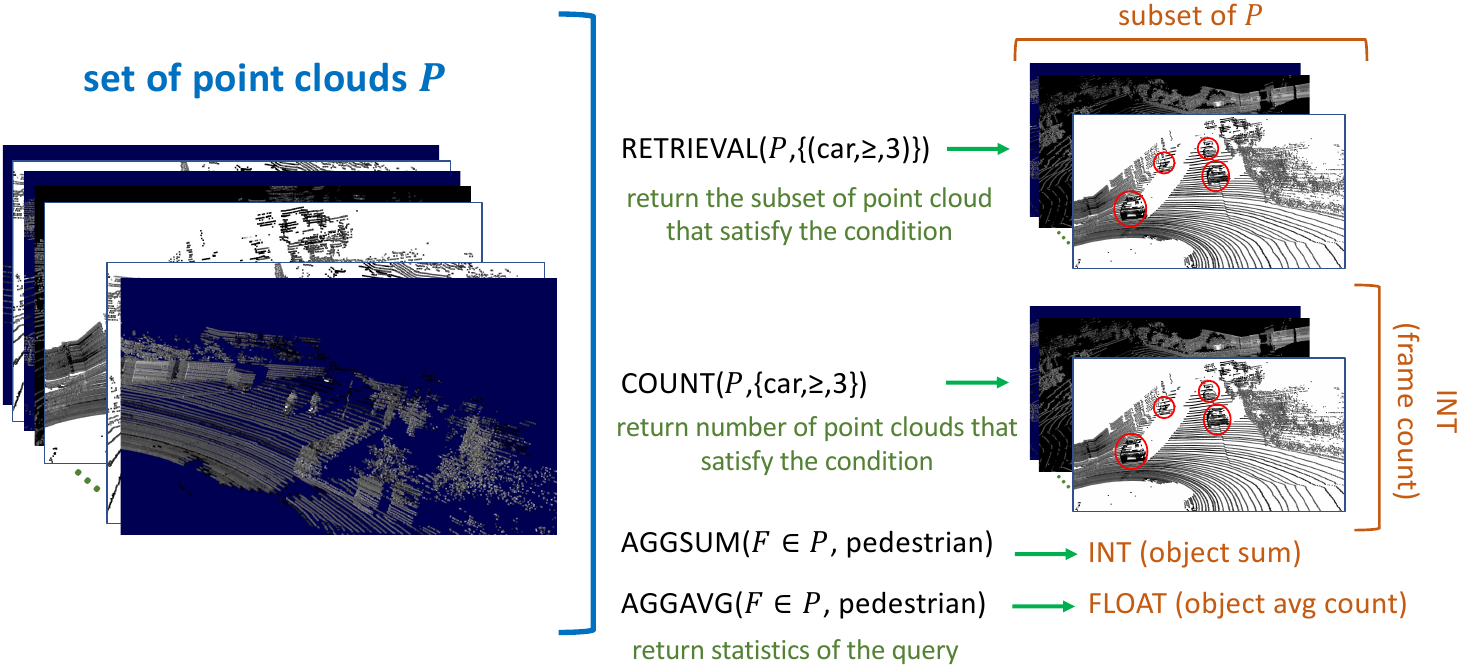}
    \centering
    \caption{Example of different query types}
    \label{fig: query_example}
\end{figure}

To support this, we define a query model that captures common patterns of analytical interest.
Drawing on prior work in traffic video and point cloud querying, we identify three key query types:
RETRIEVAL~\cite{kang13blazeit,kang2022tasti,li2025mast}, COUNT~\cite{cao2022figo}, and AGGREGATION~\cite{bang2023seiden}. These cover a broad range of analysis needs for point cloud data, as illustrated in Figure~\ref{fig: query_example}. Specifically, RETRIEVAL queries extract frames that satisfy specific conditions, useful for tasks such as object tracking and scene filtering. COUNT queries quickly quantify how many frames meet a given condition, enabling statistical analysis, monitoring, and quality control. 
AGGREGATION queries summarize patterns across multiple frames, supporting trend analysis.

\begin{figure}[t]
    \includegraphics[scale=0.2]{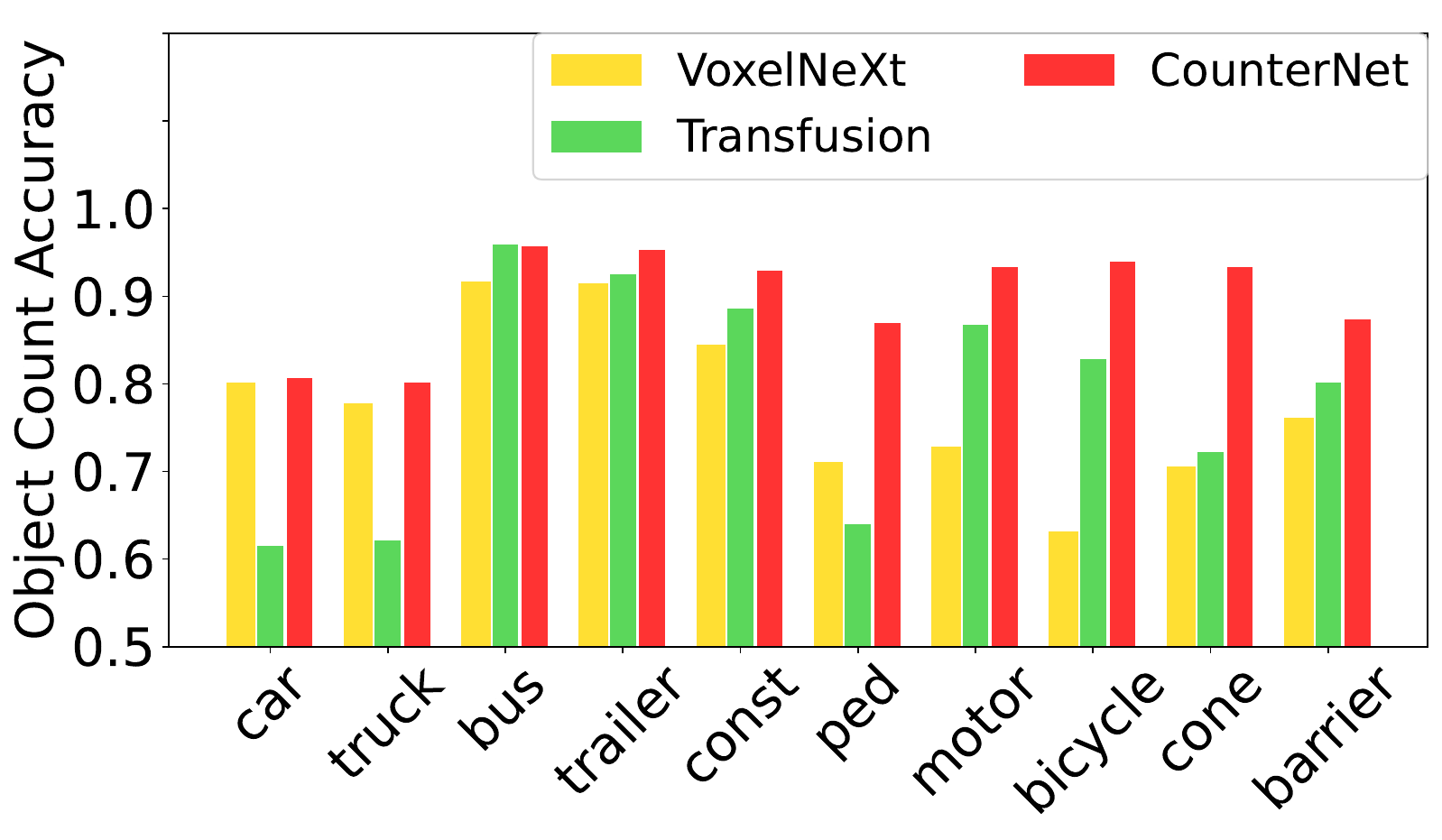}
    \centering
    \caption{Counting ACC of baselines v.s. CounterNet (ours)}
    \vspace{-.5em}
    \label{fig: baselines}
    \vspace{-.5em}
\end{figure}

These query types form the foundation of our approach. However, their effectiveness relies heavily on accurate object counts within individual frames. Unfortunately, current 3D detection models struggle to provide reliable counts across object categories, as shown in Figure~\ref{fig: baselines}. This undermines the assumption made by many existing query systems that detection outputs can serve as ground truth.

Prior work mostly focused on 2D traffic video querying\cite{kang2022tasti,kang13blazeit,bang2023seiden,cao2022figo}. These studies generally assume that 2D object detection provides ground truth semantic information, and focus primarily on building efficient and effective indexing mechanisms to reduce the cost of query execution. For example, BlazeIt~\cite{kang13blazeit} and TASTI~\cite{kang2022tasti} used proxy models to filter simple scenes, Seiden~\cite{bang2023seiden} used sampling to minimize model invocations, and Figo~\cite{cao2022figo} dynamically selected detectors based on video complexity. Although fewer works address 3D point cloud querying, approaches like MAST~\cite{li2025mast} extended Seiden~\cite{bang2023seiden} to the 3D domain by incorporating the spatial-temporal filters into the indexing process. These methods made a key assumption from their 2D counterparts: that object detection models yield ground truth results to serve as a foundation for query execution.

However, this assumption does not hold in the 3D point cloud setting. As shown in Figure~\ref{fig: baselines}, SOTA 3D detection models such as VoxelNext~\cite{chen2023voxelnext} and TransFusion~\cite{bai2022transfusion} exhibit significant variance in object counting accuracy across different classes of objects. As a result, existing 3D query systems, which rely on indexing over potentially inaccurate outputs, struggle to deliver trustworthy results. 
Unlike these indexing-focused approaches, our work takes a different path: we focus on improving object counting accuracy in point cloud data, a foundational requirement for reliable query execution. Notably, rather than improving indexing itself, we target a more fundamental problem that current methods tend to overlook.

We propose CounterNet, a heatmap-based baseline designed for accurate object counting in complex traffic scenes. Our approach focuses on enhancing semantic information extraction from point cloud frames, with a particular emphasis on achieving precise, category-aware object counts. To improve its adaptability across object types and sizes, we refine CounterNet to support multi-class counting through scale-aware mechanisms. In addition, we introduce a per-frame dynamic model selection strategy, which assigns the most suitable counting model configuration to each frame based on its complexity, further boosting overall query accuracy without sacrificing efficiency. It is worth noting that our study's primary focus is query accuracy, rather than efficiency.

\noindent\textbf{Our contributions} are summarised as below:
\begin{itemize}[leftmargin=*]
    \item To support query accuracy, we introduce CounterNet, a heatmap-based network that serves as our baseline solution. CounterNet focuses on detecting object centers, enabling object counting within large-scale traffic point cloud frames
    (Section~\ref{sec: counternet}). 
    \item We enhance CounterNet with a feature map partitioning strategy to improve performance in dense object counting (Section~\ref{sec: counternet_partition}). To better handle large objects, we introduce overlapping partitions, providing a lightweight and effective solution (Section~\ref{sec: counternet_overlap}). Finally, we propose per-frame dynamic model selection, which adaptively selects the most suitable model trained with different partition numbers and overlap ratios for each frame (Section~\ref{sec: counternet_model_selection}).
    \item We evaluated our approach on three autonomous vehicle datasets across three query types. The results show that CounterNet significantly outperforms 3D object detection models in query accuracy. Adding partitioning to CounterNet notably improved the detection of small objects, particularly in high-density scenes. Introducing overlaps between partitions further boosted performance across all object categories (Section~\ref{exp}).
\end{itemize}

\vspace{-.5em}
\section{Related Work}
\label{sec: related_work}
We review the literature from the most relevant domains: video and point cloud data querying and indexing. We further extend it to less related areas, such as 3D object detection and object counting.

\noindent\textbf{Video Data Querying and Indexing.}
Most of the related studies have focused on video data querying. Benefiting from advances in object detection in images, such as ResNet~\cite{he2016deep}, Mask R-CNN~\cite{he2017mask}, and YOLO~\cite{yolov5}, recent studies on video querying have primarily focused on indexing techniques to enhance query efficiency. For example, studies~\cite{kang2022tasti,kang13blazeit,cao2022figo} utilised proxy models as the index for efficient querying. To be specific, \citet{kang13blazeit} introduced a specialized neural network as a similarity model, employing a compact ResNet architecture. \citet{kang2022tasti} proposed a low-cost embedding deep neural network (DNN) that was trained using triplet loss. \citet{cao2022figo} focused on the ensemble of various models applied to different video segments, with the primary task being the strategic segmentation of video frames and matching segments with appropriate query models. 
In the most recent work, SEIDEN~\cite{bang2023seiden} constructed the index by effectively sampling the most important frames, allowing only these selected frames to be processed by the detection model for semantic information extraction, considering the high cost of running detection models. Additionally, SEIDEN leveraged object information extracted by the detection model to linearly predict the contents of non-sampled frames, enabling indexing for all frames. 

While these video-based techniques have demonstrated strong performance in their domain, their assumptions and design choices limit their applicability to 3D point cloud data. Specifically, point cloud data presents two challenges that are neglected in these studies: (1) Point cloud data is considerably more complex than image data~\cite{qi2017pointnet}, making it difficult to achieve effective information extraction with proxy models with simple regression. (2) Detection models for point clouds do not perform as well as those for images, meaning that even with a detection model, there is no guarantee of query accuracy in point clouds.

\noindent\textbf{Point Cloud Data Querying and Indexing.} Building on SEIDEN’s framework, \citet{li2025mast} extended this approach to traffic point cloud data. It further enhanced sampling by incorporating reinforcement learning with a spatially informed reward function and predicted the contents of non-sampled frames using spatial-temporal information extracted from sampled frames via the detection model. Similar to video query solutions, these studies primarily focused on improving query efficiency through optimized indexing. However, they often overlooked the detection model’s actual performance in extracting semantic information from point cloud frames.

\noindent\textbf{3D Object Detection in Point Clouds.} 3D object detection employs a backbone for feature extraction and a detection head for task-specific outputs. Early works such as PointNet~\cite{qi2017pointnet}, PointNet++~\cite{qi2017pointnet++}, and VoxelNet~\cite{zhou2018voxelnet} laid the foundation for 3D feature extraction. Later advancements~\cite{liu2021pvnas,zhou2020cylinder3d,lai2023spherical} introduced improved feature extraction methods. Detection heads in 3D systems follow two primary approaches: anchor-based~\cite{yan2018second,lang2019pointpillars,shi2020pv,yang20203dssd,chen2017multi} and center-based~\cite{centernet,chen2023voxelnext,bai2022transfusion,sunT0LXLA22}.
Despite substantial progress, these detection models were not designed for querying tasks. Detection focused on object identification, whereas querying emphasized extracting task-specific information.

\noindent\textbf{Object Counting in Images and Point Cloud.}
The object counts within video frames is crucial for supporting queries. Numerous studies have addressed object counting~\cite{chattopadhyay2017counting,stahl2018divide,cholakkal2020towards,laradji2018blobs,xu2023zero}, noting that counting becomes significantly harder as the number of objects increases. Recent research utilized textual information to improve accuracy~\cite{jiang2023clip,kang2024vlcounter}.
However, 3D data presented greater challenges than 2D due to the irregular distribution of points in point cloud frames, unlike the fixed grids of 2D images. While some studies investigated 3D object counting~\cite{jenkins2023countnet3d,lesani2020development,nellithimaru2019rols}, they often focused on simpler scenarios like pedestrian or crop counting. These methods struggled with the complexity of autonomous vehicle scenes, which feature diverse object types in a single frame.

\section{Problem Definition}
\label{sec: overview}
 
Autonomous vehicles generate extensive point cloud data, which is crucial for environmental understanding. Since onboard systems prioritize real-time processing, they usually have limited storage.
Instead, point cloud data is transmitted to a remote, e.g., edge server, enabling centralized processing and analysis\cite{liu2019edge,tang2021vecframe,ming2023exploration}.
Following Liu et al.~\cite{liu2019edge}, the data collected by the vehicle is uploaded to the cloud, and a NoSQL database is used to store the extracted features for querying. To achieve this, we first define the query types.

\noindent\textbf{Preliminary.} We denote $\mathcal{P}$ as a set of point cloud frames, where each point cloud frame $p \in \mathcal{P}$ is a vector in $\mathbb{R}^{3}$, representing the coordinates of a point in three-dimensional space.

\noindent\textbf{Query Types.} We define the three query types: RETRIEVAL, COUNT, and AGGREGATION. Figure~\ref{fig: query_example} demonstrates an example of each query before our formal definitions.

\begin{definition}[RETRIEVAL Query]
    The RETRIEVAL Query extracts point cloud frames from $\mathcal{P}$ where multiple objects ${q_1, q_2, ...,q_n}$ appear based on specific count constraints ${ct_1, ct_2, ...,ct_n}$. Each object has a comparison operator $op \in (\leq, \geq, =)$ to specify how its count is evaluated.
    \small

    $RETRIEVAL(\mathcal{P}, \{(q_{1}, op, ct_{1}), (q_{2}, op, ct_{2}), ..., (q_{n}, op, ct_{n})\})$
    
    $=\{p \in \mathcal{P}|\forall_{i}, \sum_{o\in p.objs} \mathbb{I}(o=q_i) \ op\ ct_i\}$

    \normalsize
   \noindent where $\mathbb{I}(\cdot)$ is an indicator function that evaluates to 1 if the condition is true, and 0 otherwise.
\end{definition}

\begin{definition}[COUNT Query]
    The COUNT Query returns the number of frames in $\mathcal{P}$ where a specified object $q$ appears, satisfying the count condition $(op, ct)$:

    \small
    $COUNT(\mathcal{P}, q, op, ct) = |\{p \in \mathcal{P}| \sum_{o \in p.objs} \mathbb{I}(o=q) \ op\ ct \}|$
\end{definition}

\begin{definition}[AGGREGATION Query]
    The AGGREGATION Query provides statistical insights over $\mathcal{P}$ for a specified object $q$. We define two aggregation functions: sum and average.
    
    $AGG_{SUM}(\mathcal{P}, q) = \sum_{p \in \mathcal{P}}\sum_{o \in p.objs} \mathbb{I} (o=q)$

    $AGG_{AVG}(\mathcal{P}, q) = \frac{1}{|\mathcal{P}|}\sum_{p \in \mathcal{P}}\sum_{o \in p.objs} \mathbb{I} (o=q)$
\end{definition}

\section{Method: CounterNet}
\label{sec: counternet}

\begin{figure*}[t]
    \includegraphics[scale=0.42]{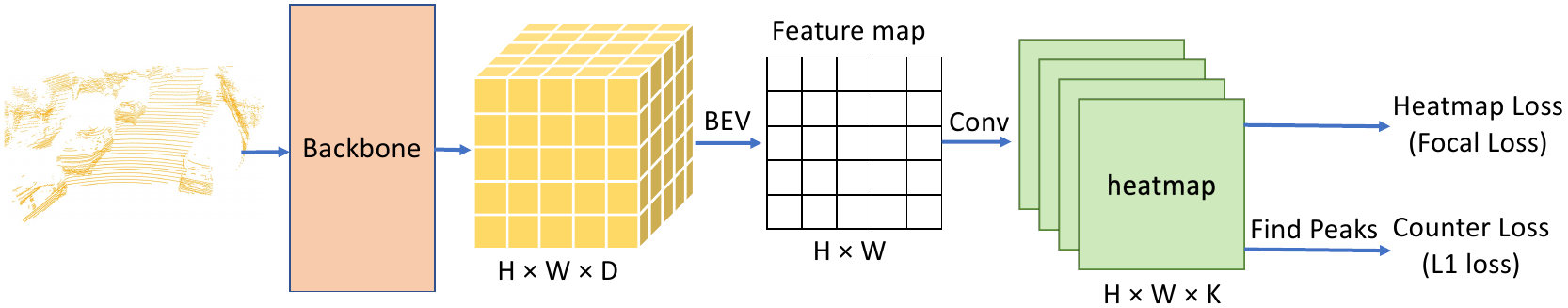}
    \centering
    \caption{CounterNet Architecture. Given a point cloud frame, the features extracted from the backbone (e.g., VoxelNet) are projected onto a 2D feature map with a bird's eye view (BEV) projection via z-axis. Subsequently, a heatmap is generated for each of the $K$ object categories based on the feature map. We use focal loss to ensure effective heatmap generation. We identify peaks within the heatmaps to get counter loss, which supervises the network to predict accurate object counts.}
    \label{fig: counternet}
\end{figure*}

\begin{figure}[t]
    \includegraphics[scale=0.25]{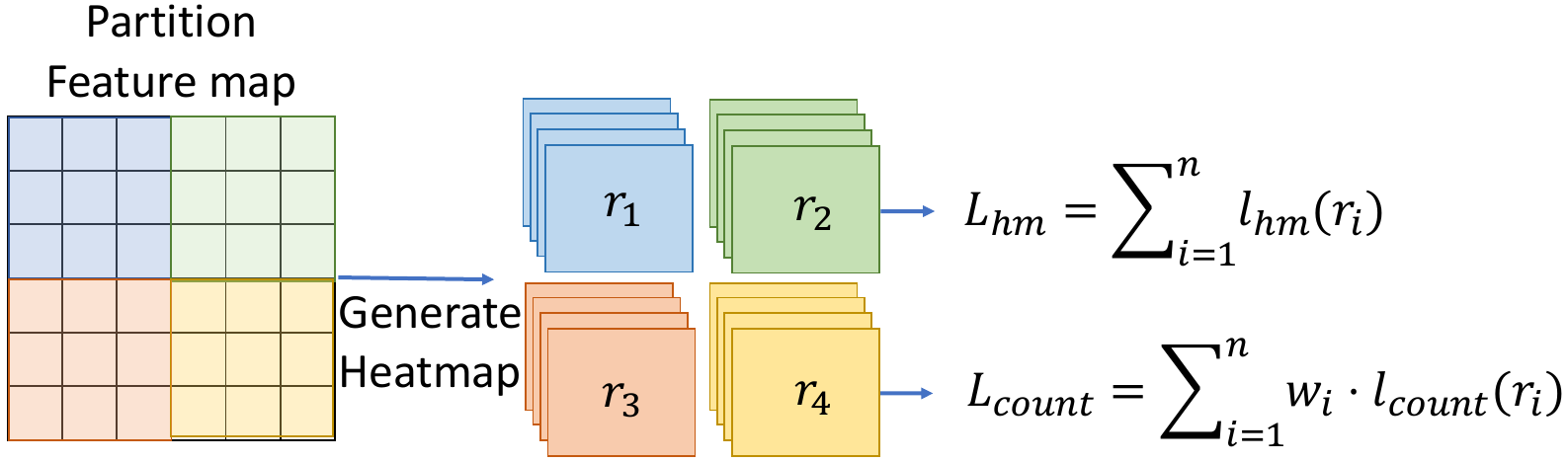}
    \centering
    \caption{Feature Map Partition}
    \vspace{-.8em}
    \label{fig: counternet_partition}
\end{figure}

\textbf{Data Model.} Given that object count is the most important information for querying~\cite{kang2022tasti,kang13blazeit,cao2022figo,bang2023seiden}, we define a NoSQL data model as the document-like structure. The detailed definition can be found in Appendix~\ref{app: data_model}.

\noindent\textbf{Design Motivation of CounterNet.} The most straightforward method for obtaining object counts is through a regression model. However, a single point cloud frame contains vast amounts of information, including multiple objects in varying numbers, making it difficult for a regression model to capture such fine-grained details. Alternatively, object detection models are focused on localizing objects, such as drawing accurate bounding boxes and determining object orientation. This complexity causes them to underperform in tasks focused solely on object counting.


This motivates us to develop a method that has the following properties: (1) To mitigate the limitation in the regression model, the method should be capable of detecting the appearance of the object in the region of the point cloud for a more fine-grained detection. (2) To mitigate the limitation brought by the detection model, the method should concentrate on detecting the appearance of objects instead of spending efforts on the localization of objects.

\noindent\textbf{Network Architecture.} 
We propose to utilise heatmaps for object counting, as illustrated in Figure~\ref{fig: counternet}.
The objective of this approach is to detect object centers, which are represented as peaks in the heatmap. This method satisfies the two expected property mentioned above: (1) Compared to simple regression, the heatmap emphasizes identifying object appearances by detecting centers, resulting in more accurate counts. (2) It simplifies the detection task by disregarding object size and orientation, focusing exclusively on detecting the object centers to identify their presence.

\noindent\textbf{Heatmap.} With the projected feature map, we generate a K-channel heatmap ($K = |C|$), where each channel corresponds to a specific object type. To train the heatmap, we utilise the ground truth positions of objects, annotating object centers with higher values and surrounding areas with lower values~\cite{centernet,law2018cornernet,yin2021centerpoint}. 

\noindent\textbf{Object Counting via Threshold-based Peak Detection.} 
We count objects by finding local peaks in the heatmap. These peaks correspond to detected object centers. Specifically, we apply a 2D local maxima algorithm \footnote{\url{https://mathworks.com/help/matlab/ref/islocalmax2.html}} to identify object centers, which then allows us to estimate both the count and approximate positions of the objects. Thresholding is crucial for reducing noise. 
To suppress noise in low-activation regions of heatmaps, we apply a fixed threshold to the predicted heatmap before applying local maxima detection. This ensures that only confident peaks, corresponding to likely object centers, are retained. The threshold is determined during the training stage. Since the heatmap values are normalized within the range of 0 to 1, we select a mid-range value as the threshold to ensure effective filtering (i.e., 0.5). Intuitively, the model is expected to assign higher values to regions corresponding to object grids, while non-object regions will have lower values.

\noindent\textbf{Loss.}
As shown in Figure~\ref{fig: counternet}, we use two losses, focal loss and counter loss, to supervise the network for accurate counting. 

We use the focal loss \cite{lin2017focal} to supervise the generation of the heatmap, as defined in Equation~\ref{eq: focal_loss}. In this equation, $\hat{Y}_{xyc}$ represents the predicted probability at location $(x, y)$ in the heatmap, while $Y_{xyc}$ denotes the ground truth label for object $c$, with $Y_{xyc} = 1$ indicating the presence of an object center. The term $N$ refers to the number of positive samples (i.e., locations where $Y_{xyc} = 1$), which is used to normalize the loss. 
The parameter $\alpha$ is a focusing factor. We take the value $\alpha=2$ by following~\cite{centernet}.
\begin{equation}
\small
    L_{hm}=\frac{-1}{N} \sum_{\hat{Y}}\left\{\begin{array}{l}\left(1-\hat{Y}_{x y c}\right)^\alpha \log \left(\hat{Y}_{x y c}\right) \quad \text { if } Y_{x y c}=1 \\ \left(1 - Y_{x y c}\right)^\alpha \log \left(1-\hat{Y}_{x y c}\right) \text { otherwise }\end{array}\right.
    \label{eq: focal_loss}
\end{equation}

We use thresholded local maxima on the heatmap to derive the object count. To supervise this process, we introduce an L1 loss for counting accuracy as shown in Equation~\ref{eq: l1_loss}. In this equation, $y$ represents the predicted count, while $\hat{y}$ represents the ground truth count. The count loss provides an additional supervisory signal to the heatmap, ensuring that regions corresponding to object centers exceed the threshold while other regions remain below it. 

\begin{equation}
    L_{count} = \frac{1}{|C|} \frac{1}{|\mathcal{P}|} \sum^{|C|}_{j=1} \sum^{|\mathcal{P}|}_{i=1} |y_{ji} - \hat{y}_{ji}|
     \label{eq: l1_loss}
\end{equation}

We combine two losses: heatmap loss and counter loss for the supervision of the network $\mathcal{L} = L_{hm} + L_{count}$.

\section{Performance Optimizations of CounterNet}
 To further enhance CounterNet's performance, we introduce a partition-based approach that divides the point cloud into smaller regions, improving counting accuracy (in Section~\ref{sec: counternet_partition}). Additionally, we propose adding overlap between partitions to more accurately detect objects near the partition boundaries, leading to more precise and reliable counting results (in Section~\ref{sec: counternet_overlap}). We also provide a model selection method to dynamically select the model trained with different parameters to further optimize the counting accuracy across different object types (in Section~\ref{sec: counternet_model_selection}).
\subsection{CounterNet with Feature Map Partition}
\label{sec: counternet_partition}

Although CounterNet performs well overall, it struggles when there are dense objects in a single frame, which is consistent with the previous findings in images that counting a higher number of objects is inherently more difficult than counting fewer~\cite{stahl2018divide,chattopadhyay2017counting}.
Additionally, one of the reasons for this difficulty is the imbalance in the training dataset: frames with many objects are underrepresented, while most frames contain only a few or no objects at all. This imbalance makes it harder for the model to generalize well to scenes with a higher object count.
To address this issue and improve CounterNet's performance in cases where dense objects are present, we propose CounterNet with partitioning. This method is designed to handle high object density cases more effectively, allowing for more accurate counting in complex frames.

\noindent\textbf{Network Architecture. } 
We evenly partition the feature map into regions of equal size. We denote the number of partitions as $pt$. For example, Figure~\ref{fig: counternet_partition} illustrates the case where $pt=4$, resulting in four equally sized regions within the feature map. Each region is then passed through convolutional layers to produce a partial heatmap corresponding to that region.

This approach offers several advantages. First, by breaking the feature map into smaller regions, we reduce the object number in each partition. It lowers the complexity of detecting and counting objects within partitions. Second, partitioning increases the amount of effective training data by generating more object-containing regions. By assigning more weight to the loss in data-rich regions, we improve the precision of heatmap generation, ultimately enhancing the overall accuracy of counting.

\noindent\textbf{Dynamic Thresholding for Peak Detection.} 
Once a heatmap is generated from the partitions, using a single threshold for all partitions no longer yields optimal results. For instance, in a large heatmap, high-density partitions tend to have higher overall values, while low-density partitions have lower values. A single threshold applied to all partitions cannot account for these variations, resulting in suboptimal performance.

To address this, we apply Otsu’s method~\cite{otsu1975threshold} during the inference stage to dynamically determine the optimal threshold $k$ for each partition. Otsu’s method maximizes the between-class variance, given by the formula: $\alpha_B^2(k)=\omega_0(k) \times \omega_1(k) \times\left[mean_0(k)-mean_1(k)\right]^2$. Here, $\omega_0(k)$ and $\omega_1(k)$ are the proportion of points below and above the threshold, respectively, while $mean_0(k)$ and $mean_1(k)$ represent the mean values of the points below and above $k$, respectively. By maximizing this function, we find the optimal threshold $k$ for each partition.
The computed threshold $k$ is then used to distinguish relevant and irrelevant regions within the partition. 
We combine this dynamic threshold $k$ with the fixed threshold $t$ in training. If $k \geq t$, we adopt $k$ as the threshold for partition; otherwise, we use $t$ as the threshold. 

\noindent\textbf{Loss. }
Partitioning the feature map introduces imbalance across partitions. To mitigate this, we apply weights to the count loss for each partition, as shown in Equation~\ref{eq: loss_partition}, allowing the model to focus more on high-density partitions.
\begin{equation}
\small
    L_{count} = \sum_{i=1}^{|R|}w_i \cdot l_{count} (r_i),w_i = \frac{1}{|R|} + \frac{|r_i.objs|}{\sum_{i=1}^{|R|}|r_i.objs|}
    \label{eq: loss_partition}
\end{equation}

Our weight has two parts: the base weight and the counter weight. The base weight, $\frac{1}{|R|}$, is uniformly assigned to all partitions, where $|R|$ is the total number of partitions. The counter weight, $\frac{|r_i.objs|}{\sum_{i=1}^{|R|}|r_i.objs|}$, is calculated according to the number of objects in each partition. Here, $|r_i.objs|$ is the number of objects in the partition, while $\sum_{i=1}^{|R|}|r_i.objs|$ is the total objects across all partitions.

\subsection{Feature Map Partition with Overlaps}
\label{sec: counternet_overlap}

\begin{figure}[t]
    \includegraphics[scale=0.18]{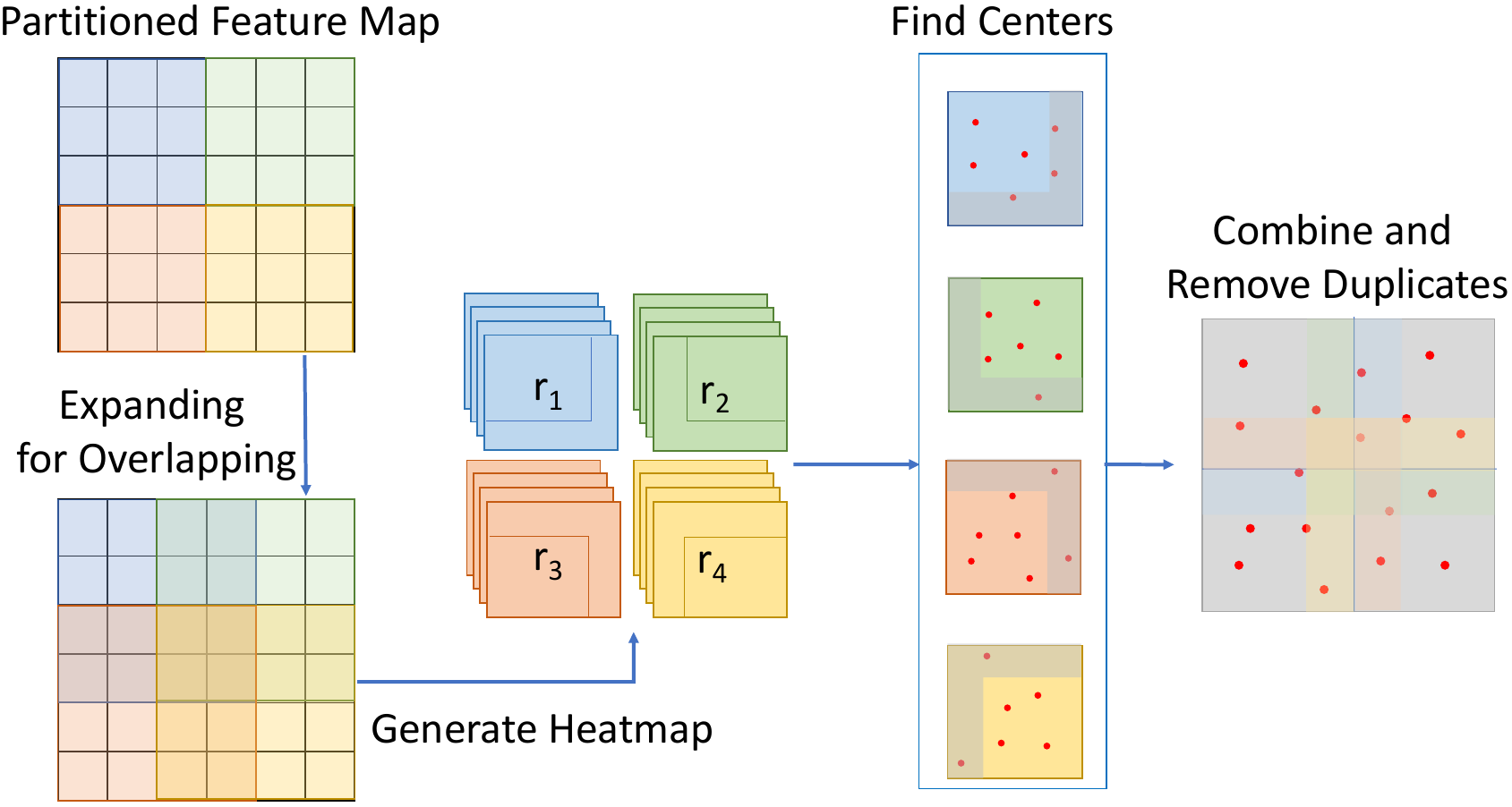}
    \centering
    \caption{Feature Map Partition with Overlap}
    \label{fig: partition_overlap}
\end{figure}

While partitioning improves counting accuracy in high-density frames, we find that it can sometimes split a single object across multiple partitions, leading to the loss of tracking for that object. This raises a key question: how do we handle cases where an object's center falls near the boundary of a partition?

An intuitive approach to prevent objects from being split across partitions is to use feature-based clustering methods such as K-Means~\cite{faber1994kmeans}. However, clustering presents several challenges in this context. First, it is computationally expensive and struggles to differentiate objects where object boundaries may be unclear. Second, clustering
produces irregular regions that disrupt backpropagation during training. These limitations make clustering an impractical solution for integrating with our pipelines.

To overcome these challenges, we propose a simpler and more effective alternative by expanding partition regions to introduce overlap, as illustrated in Figure~\ref{fig: partition_overlap}. This approach reduces the likelihood of splitting an object across partitions by ensuring that if an object's center falls near a partition boundary, it is also captured within the central region of an adjacent partition. Compared to clustering, this method has several advantages: (1) it is lightweight and adds minimal computational overhead while preserving the benefits of partitioning, and (2) it integrates seamlessly into the CounterNet pipeline, maintaining consistency across frames.

The training and inference details is in Appendix~\ref{app: overlap_partition}.

\subsection{Query Optimization by Model Selection}
\label{sec: counternet_model_selection}

\begin{figure}[t]
    \includegraphics[scale=0.25]{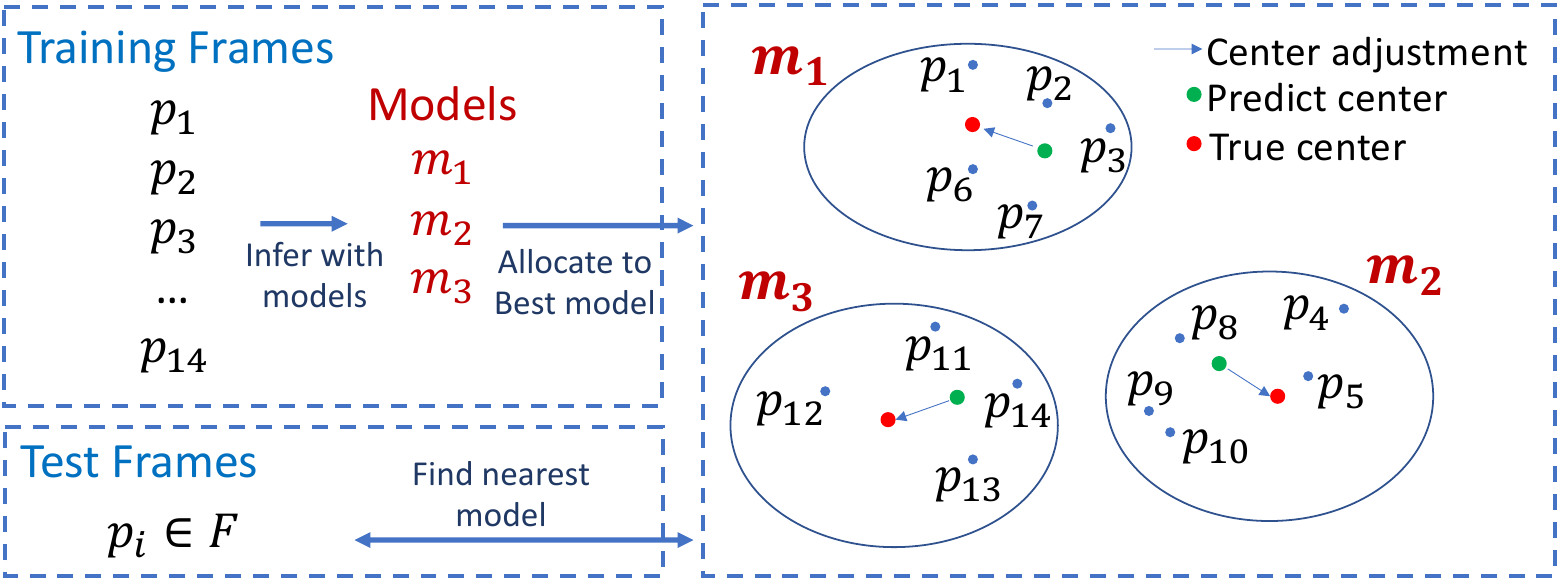}
    \centering
    \caption{Process of Model Selection}
    \label{fig: model_selection}
\end{figure}

Autonomous vehicle point cloud frames vary significantly, making object counting challenging for models with fixed partitions and overlap ratios. Small, numerous objects benefit from more partitions, while larger objects require fewer. We propose a model selection method that chooses the nearest model based on feature distance. To mitigate errors in model center estimation, we apply the Chernoff bound to adjust selection probabilistically. The algorithm can be found in Appendix~\ref{app: model_selection}.

\noindent\textbf{Model Center Representation.}  
Given a set of models, denoted as $m \in M$, each corresponding to different partition numbers and overlap ratios, we perform inference on every point cloud frame $p$ in the training dataset using all models in $M$. For each point cloud frame, we identify the model that demonstrates the best performance. For instance, as shown in Figure~\ref{fig: model_selection}, the training point cloud frames are allocated to the models as follows: $m_1$ performs best on $p_1, p_2, p_3, p_6, p_7$, $m_2$ performs best on $p_4, p_5, p_8, p_9, p_{10}$, and $m_3$ performs best on $p_{11}, p_{12}, p_{13}, p_{14}$.
    
For each model $m_j$, we estimate its center feature, represented as $\hat{\omega}_{m_j}$, by calculating the average feature vector across all point cloud frames associated with the model. The center estimate is computed as:
$\hat{\omega}_{m_j} = \frac{1}{n}\sum_{i=0}^{n} p_i.f$
where $p_i.f$ demoted the feature vector of point cloud frame $p_i$, and $n$ is the number of point cloud frames assiciated with model $m_j$.

\noindent\textbf{Center Adjustment with Confidence.} 
Under the assumption of unlimited data for each $m\in M$, we would ideally obtain the true center $\mu$ for each model. However, in practice, data distribution across models is often uneven, making it necessary to estimate the probability of deviation between the evaluated center $\hat{\omega}_{m_j}$ and the true center $\mu$.

To quantify the probability of deviation, we assume point cloud frames are independent and apply the Chernoff bound~\cite{hellman1970chernoff}, which provides an upper bound on the probability that $\hat{\omega}_{m_j}$ deviates from $\mu$ by more than a given margin $\epsilon$:     $P(|\hat{\omega}_{m_j} - \mu| > \epsilon)\leq exp(-\frac{n\epsilon^2}{2\xi^2})$.
Here, $\epsilon$ is the expected deviation, $\xi$ is the sample variance, and $n$ is the number of samples.
In this equation, $\epsilon$ is the distance threshold defined by the user. Lower $\epsilon$ provides a higher probability. This probability $P_{m_j}$ quantifies our confidence in the estimated center $\hat{\omega}_{m_j}$, guiding adjustments to improve model selection.


\noindent\textbf{Finding the optimal model.} During validation, for each new frame $p$, we compute the distance between $p$ and the center of each model $m \in M$, . The model with the shortest distance to the new point cloud frame is selected for further processing. Additionally, when selecting the closest model center, we must take into account the potential deviation of each model's estimated center from its true center. For the distance between $p$ to each $m \in M$, we combine the distance with the probability: $d_{adjusted}(p, m_j) = d(p, \hat{\omega}_{m_j}) \times P_{m_j}$, where $d(p, \hat{\omega}_{m_j})$ is the distance between $p$ and $\hat{\omega}_{m_j}$, and $P_{m_j}$ is the confidence probability derived from the Chernoff bound. By incorporating $P_{m_j}$, we adjust for potential deviation in the estimated center and improve robustness in model selection.

\section{Experiment}
\label{exp}

\subsection{Experiment Setup}
All experiments were conducted on a server running
Ubuntu 22.04.1, with Intel(R) Xeon(R) CPU E5-2697A v4
@ 2.60GHz and RTX3080. The code was implemented with OpenPCDet Framework~\cite{openpcdet2020}. Our code is available at~\cite{mygit}.

\noindent\textbf{Datasets.} Experiments were conducted on the nuScenes~\cite{nuscenes2019}, KITTI~\cite{kitti2019iccv} and Waymo-partial~\cite{waymoPerception2020} (detailed explanation in Appendix~\ref{app: waymo}). 

\begin{table}[h]
\small
\centering
\begin{tabular}{|l|p{6cm}|}
\hline
\textbf{Notation} & \textbf{Description} \\ \hline
\( CN \) & CounterNet baseline without partitioning. \\ \hline
\( CN_{\text{pt}} \) & CounterNet with partitioning. The default setting uses 4 partitions unless otherwise specified. \\ \hline
\( CN_{\text{pt}}^{o} \) & CounterNet with partitioning and overlap. Uses the expansion ratio of 0.2 by default. \\ \hline
\end{tabular}
\caption{Notations in experiments}
\label{table: cn_notation}
\vspace{-1cm}
\end{table}

\begin{table*}[t!]
\small
\begin{tabular}{|l|l|ccc|ccc|ccc|ccc|}
\hline
                   &       & \multicolumn{3}{c|}{RETRIEVAL-S (10\% tolerance)} & \multicolumn{3}{c|}{RETRIEVAL-S (5\% tolerance)} & \multicolumn{3}{c|}{AGG (absolute)} & \multicolumn{3}{c|}{AGG (Q-error)} \\ \cline{3-14} 
                   &       & nuScenes         & KITTI         & Waymo         & nuScenes         & KITTI         & Waymo         & nuScenes     & KITTI    & Waymo    & nuScenes    & KITTI    & Waymo    \\ \hline
\multirow{4}{*}{\rotatebox{90}{baselines}} & VN    & 0.829            & --            & 0.779         & 0.706            & --            & 0.671         & 2.637        & --       & 8.884    & 1.804       & --       & 1.268    \\
                   & TF    & 0.857            & --            & --            & 0.767            & --            & --            & 2.706        & --       & --       & 1.774       & --       & --       \\
                   & SN    & 0.871            & --            & --            & 0.794            & --            & --            & 1.778        & --       & --       & 1.556        & --       & --       \\
                   & CP    & --               & 0.917         & --            & --               & 0.828         & --            & --           & 0.330    & --       & --          & 1.286    & --       \\ \hline
\multirow{3}{*}{\rotatebox{90}{ours}} & $CN$    & 0.901            & 0.939         & 0.812         & 0.823            & 0.846         & 0.711         & 1.228        & \textbf{0.189}    & 6.391    & 1.416       & 1.080    & 1.248    \\
                   & $CN_{pt}$  & 0.878$^\downarrow$  & 0.935$^\downarrow$  & 0.818  & 0.829    & 0.840$^\downarrow$  & 0.630$^\downarrow$  & 1.053  & 2.501$^\uparrow$    & 3.560    & 1.428$^\uparrow$  & 1.067    & 1.199    \\
                   & $CN_{pt}^o$ & \textbf{0.916}  & \textbf{0.954} & \textbf{0.846} & \textbf{0.839} & \textbf{0.859} & \textbf{0.712} & \textbf{0.744}  & 0.276  & \textbf{3.301} & \textbf{1.321} & \textbf{1.051} & \textbf{1.197}    \\ \hline
\end{tabular}

\caption*{\footnotesize $\downarrow$ indicates a performance drop for RETRIEVAL, while $\uparrow$ means a performance drop for AGG, compared to $CN$. Bold indicates the best performance.}
\vspace{-1em}
\captionof{table}{Overall query result of RETRIEVAL-SINGLE and AGGREGATION}
\vspace{-1.5em}
\label{table: all_retrieval_agg}
\end{table*}
\begin{figure*}
    \tiny
    \begin{minipage}{0.49\textwidth}
\small
\centering
\begin{tabular}{|lp{0.4cm}p{0.4cm}p{0.4cm}p{0.4cm}p{0.7cm}p{0.7cm}|}
\hline
 Combinations                                                                   & VN    & TF  & SN  & $CN$    & $CN_{pt}$ & $CN_{pt}^{o}$ \\ \hline
(car, pedestrian)                                                     & 0.582 & 0.425 & 0.675 & 0.700 & 0.800  & 0.842   \\
(car, barrier)                                                        & 0.616 & 0.506 & 0.698 & 0.693 & 0.808  & 0.852   \\
\begin{tabular}[c]{@{}l@{}}(pedestrian, \\ barrier)\end{tabular}      & 0.568 & 0.540 & 0.649 & 0.783 & 0.791$^\downarrow$  & 0.787   \\
\begin{tabular}[c]{@{}l@{}}(car, pedestrian, \\ barrier)\end{tabular} & 0.474 & 0.372 & 0.571 & 0.623 & 0.724  & 0.759   \\ \hline
(truck, bus)                                                          & 0.725 & 0.605 & 0.794 & 0.771 & 0.758$^\downarrow$  & 0.828   \\
(bus, trailer)                                                        & 0.844 & 0.890 & 0.921 & 0.910 & 0.897$^\downarrow$  & 0.913   \\
(truck, trailer)                                                      & 0.732 & 0.607 & 0.784 & 0.791 & 0.773$^\downarrow$  & 0.832   \\
\begin{tabular}[c]{@{}l@{}}(truck, bus,\\ trailer)\end{tabular}       & 0.684 & 0.591 & 0.773 & 0.760 & 0.733$^\downarrow$  & 0.804   \\ \hline
\end{tabular}
\vspace{-1em}
\caption*{\footnotesize $\downarrow$ indicates a performance drop compared to $CN$.}
\captionof{table}{RETRIEVAL-MULTIPLE (nuScenes)}
\label{table: join}

    \end{minipage}
    \begin{minipage}{0.49\textwidth}
\centering
\small
\begin{tabular}{|lcccc|}
\hline
                                                                      & CP    & $CN$    & $CN_{pt}$ & $CN_{pt}^o$ \\ \hline
(car, pedestrian)                                                     & 0.812 & 0.856 & 0.854$^\downarrow$  & \textbf{0.902}   \\
(car, cyclist)                                                        & 0.885 & 0.887 & 0.880$^\downarrow$  & \textbf{0.903}   \\
(pedestrian, cyclist)                                                 & 0.833 & 0.886 & 0.868$^\downarrow$  & \textbf{0.926}   \\
\begin{tabular}[c]{@{}l@{}}(car, pedestrian, \\ cyclist)\end{tabular} & 0.776 & 0.821 & 0.811$^\downarrow$  & \textbf{0.869}   \\ \hline
\end{tabular}
\captionof{table}{ETRIEVAL-MULTIPLE (KITTI)}
\vspace{-0.15cm}
\label{table: kitti_join}

\centering
\small
\begin{tabular}{|lcccc|}
\hline
                              & VN                         & $CN$                        & $CN_{pt}$                     & $CN_{pt}^o$                      \\ \hline
(vehicle, pedestrian)         & 0.757                      & 0.798                     & 0.748$^\downarrow$                     & \textbf{0.835 }                     \\
(vehicle, cyclist)            & 0.751                      & 0.792                     & 0.733$^\downarrow$                     & \textbf{0.832 }                     \\
(pedestrian, cyclist)         & 0.757                      & 0.790                     & 0.748$^\downarrow$                     & \textbf{0.831}                      \\
(vehicle, pedestrian, cyclist) & \multicolumn{1}{l}{0.745} & \multicolumn{1}{l}{0.780} & \multicolumn{1}{l}{0.692$^\downarrow$} & \multicolumn{1}{l|}{\textbf{0.818}} \\ \hline
\end{tabular}
\captionof{table}{RETRIEVAL-MULTIPLE (Waymo)}
\label{table: waymo_join}
    \end{minipage}
\end{figure*}

\noindent\textbf{Methods for Comparison.} As explained in Section~\ref{sec: related_work}, unlike prior querying studies that focus on indexing~\cite{li2025mast,bang2023seiden,kang13blazeit,kang2022tasti}, our solution focuses on facilitating querying performance by improving object counting. As a result, we use the following \textbf{detection models as our baselines} for querying: (1) VoxelNeXt~\cite{chen2023voxelnext} (VN), which relies on sparse voxel features to facilitate detection; (2) TransFusion~\cite{bai2022transfusion} (TF), which utilizes an attention mechanism to fuse point cloud and image data for improved detection; (3) CenterPoint~\cite{yin2021centerpoint} (CP), a pioneer study of the center-based detection model;
(4) SAFDNet~\cite{safdnet} (SN), which utilises a fully sparse feature map for object detection. SAFDNet is one of the most recent detection studies, and shows better performance than other baselines.

Given the availability of baseline implementations on different datasets, we use VoxelNeXt, TransFusion and SAFDNet as baselines for the nuScenes dataset. For the Waymo dataset, we adopt VoxelNeXt, while for the KITTI dataset, we adopt CenterPoint.

\noindent We also have \textbf{our methods} for comparison:

\noindent(1) CounterNet \textbf{($CN$)}: Our proposed baseline in Section~\ref{sec: counternet}.

\noindent(2) CounterNet with partition \textbf{($CN_{pt}$)}: Our proposed solution, as introduced in Section~\ref{sec: counternet_partition}.

\noindent(3) CounterNet with overlapped partition \textbf{($CN_{pt}^{o}$)}: Our proposed solution, as introduced in Section~\ref{sec: counternet_overlap}.

\noindent\textbf{Evaluation Metrics.} We evaluate our approach based on two main aspects: query accuracy and counting performance. 

\noindent(1) RETRIEVAL. 
We experiment with two types of retrieval queries: RETRIEVAL-SINGLE and RETRIEVAL-MULTIPLE. These queries retrieve frames containing a specific number of objects. In most analytical scenarios, such as congestion detection, precise object counts are not strictly necessary; instead, a degree of deviation is acceptable to account for natural variability. To evaluate performance, we calculate the percentage of frames in which the retrieval condition was correctly met, allowing for 10\% and 5\% error tolerance. 
Specifically, the tolerance threshold for each object category is determined by multiplying the maximum number of that object type in a scene by the error tolerance. This approach ensures a fair evaluation across categories.

\noindent(2) COUNT. The COUNT query counts the number of frames that satisfy a specific condition, such as counting the frames containing 5 cars.
To evaluate this, we randomly select 500 groups of consecutive frames with lengths ranging from 100 to 500. For each object category, we generate 1000 random queries requesting different object numbers (not exceeding the maximum number of objects present in any frame) and evaluate the percentage of counts being returned correctly. We allow a 10\% error tolerance in the results.

\noindent(3) AGGREGATION.
The aggregation query provides statistical information such as total number of objects in a set of frames. We evaluate this using the SUM() operation.
For random consecutive frame groups of lengths between 100 and 500, we calculate the sum of queried objects and compare it to the ground truth across 1000 random queries. We use two metrics to evaluate the results: absolute difference and Q-error. The absolute difference measures the average deviation between the predicted and ground truth sums, providing a straightforward sense of numerical accuracy. In contrast, Q-error (quantile error) evaluates the ratio between the predicted and actual values, capturing the relative error~\cite{qerror2021}.

\noindent\textbf{Notation and Parameter Setting.} Please refer to Table~\ref{table: cn_notation} for the notation of different CounterNet variants and the default values.

\subsection{Effectiveness Study on Various Query Types}

Table~\ref{table: all_retrieval_agg} and Table~\ref{table: all_count} demonstrate the overall performance of the three query types: RETRIEVAL, AGGREGATION, and COUNT. Across all three queries, the partition-based model ($CN_{pt}$) does not consistently outperform the baseline ($CN$), with performance drops particularly obvious on the KITTI dataset in both \textit{RETRIEVAL-SINGLE (i.e., RETRIEVAL-S)} and \textit{AGGREGATION (i.e. AGG)} queries. This decline is likely due to KITTI’s smaller spatial coverage per point cloud frame, making it more sensitive to partitioning. However, introducing overlap in the partitioning strategy ($CN_{pt}^o$) effectively mitigates this issue, delivering the best performance in most cases and demonstrating greater robustness across datasets of varying scales. Although all models exhibit decreased \textit{RETRIEVAL} performance under the stricter 5\% error tolerance, the degradation is more pronounced in $CN_{pt}$, while $CN_{pt}^o$ remains comparatively stable, further highlighting its resilience.

Tables~\ref{table: join}-\ref{table: waymo_join} show RETRIEVAL-MULTIPLE results on different datasets. In Table~\ref{table: join}, we evaluate two groups based on occurrence frequency: high-frequency objects (cars, pedestrians, barriers) and low-frequency objects (trucks, buses, trailers), by referring to Table~\ref{table: nus_distribution} for object frequency on nuScenes in Appendix~\ref{app: nuscenes}. (1) High-Frequency Objects: Partitions enhance the result, and overlap further improves it, except for pedestrians and barriers, where duplicate counts in overlapped regions reduce accuracy due to their smaller size and higher occurrence rates. (2) Low-Frequency Objects: Partitions do not improve the accuracy, but overlap significantly enhances performance, particularly for sparsely distributed or large objects.
The detailed categorical results and analysis can be found in Appendix~\ref{app: nuscenes} - \ref{app: waymo}.

\subsection{Parameter Study}

\begin{figure*}
\begin{minipage}{0.34\textwidth}
    \centering
    \includegraphics[scale=0.19]{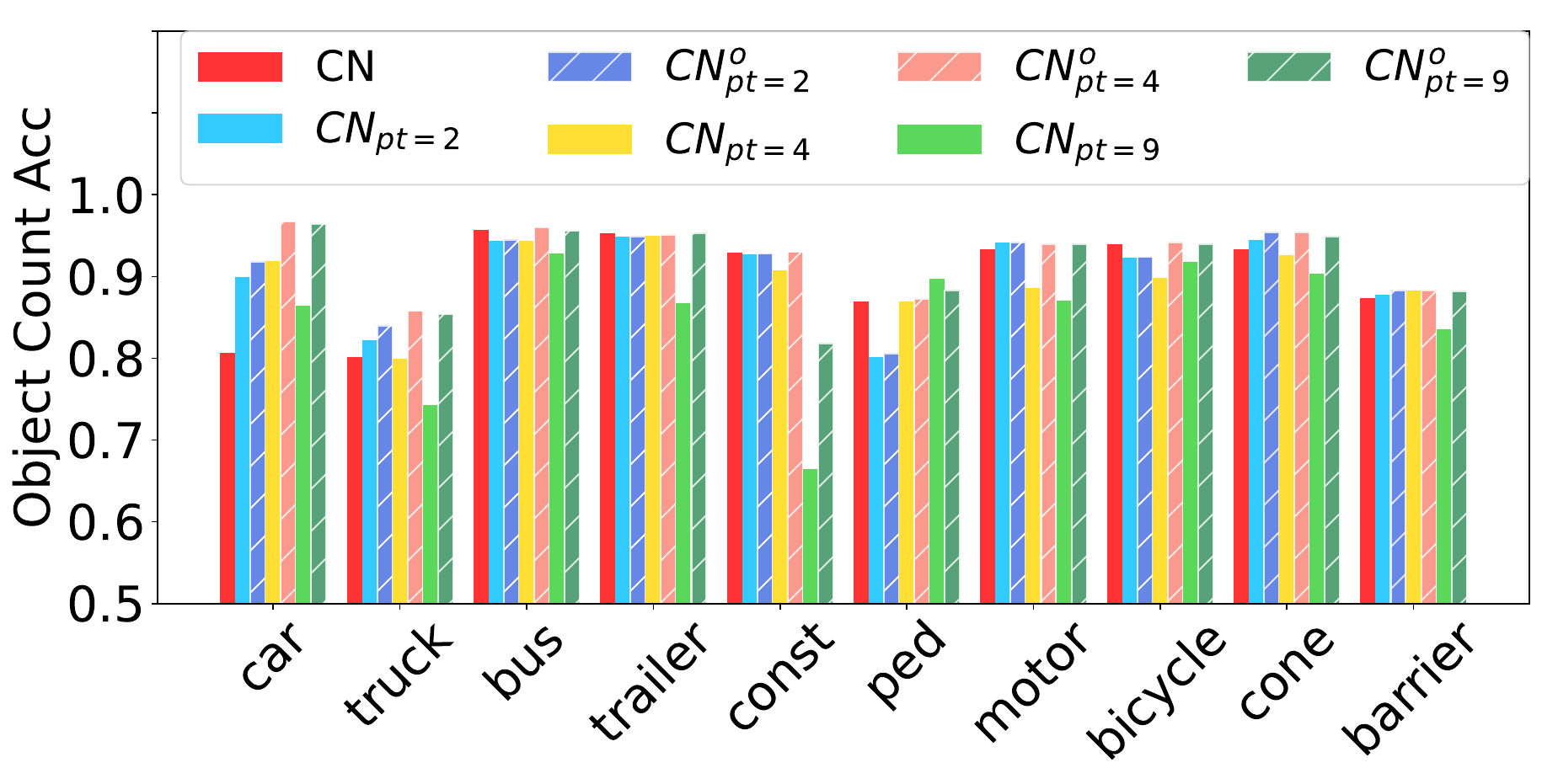}
    \vspace{-0.5em}
    \caption{Different Partitions(nuScenes)}
    \label{fig: para_partition}
\end{minipage}
\hfill 
\begin{minipage}{0.325\textwidth}
    \centering
    \includegraphics[width=\textwidth]{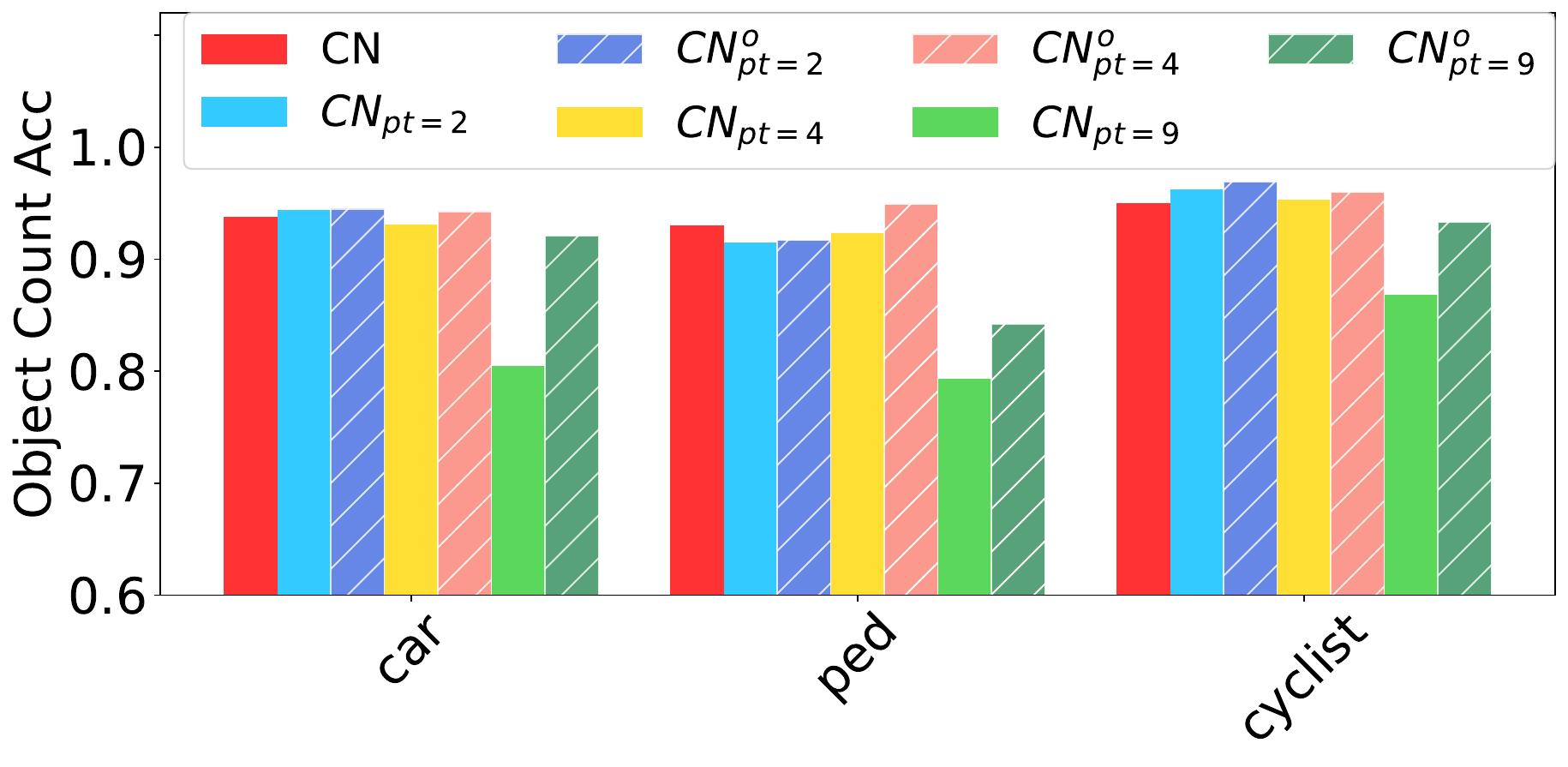}
    \caption{Different partitions(KITTI)}
    \label{fig: diff_partition_kitti}
\end{minipage}
\begin{minipage}{0.325\textwidth}
    \includegraphics[width=\textwidth]{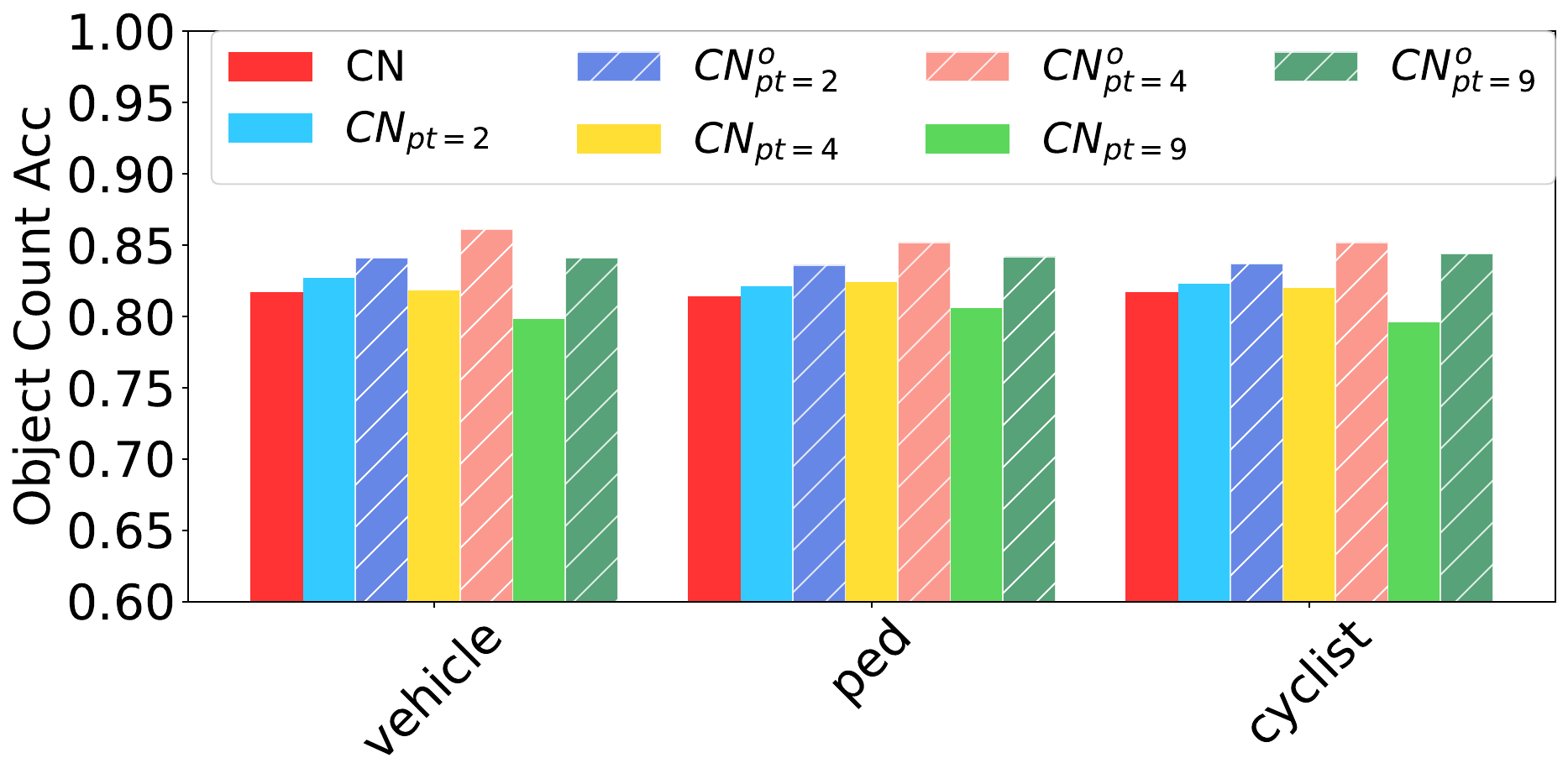}
    \centering
    \captionof{figure}{Different partitions(Waymo)}
    \label{fig: diff_partition_waymo}
\end{minipage}
\end{figure*}

\begin{figure*}
\begin{minipage}{0.34\textwidth}
    \centering
    \includegraphics[scale=0.21]{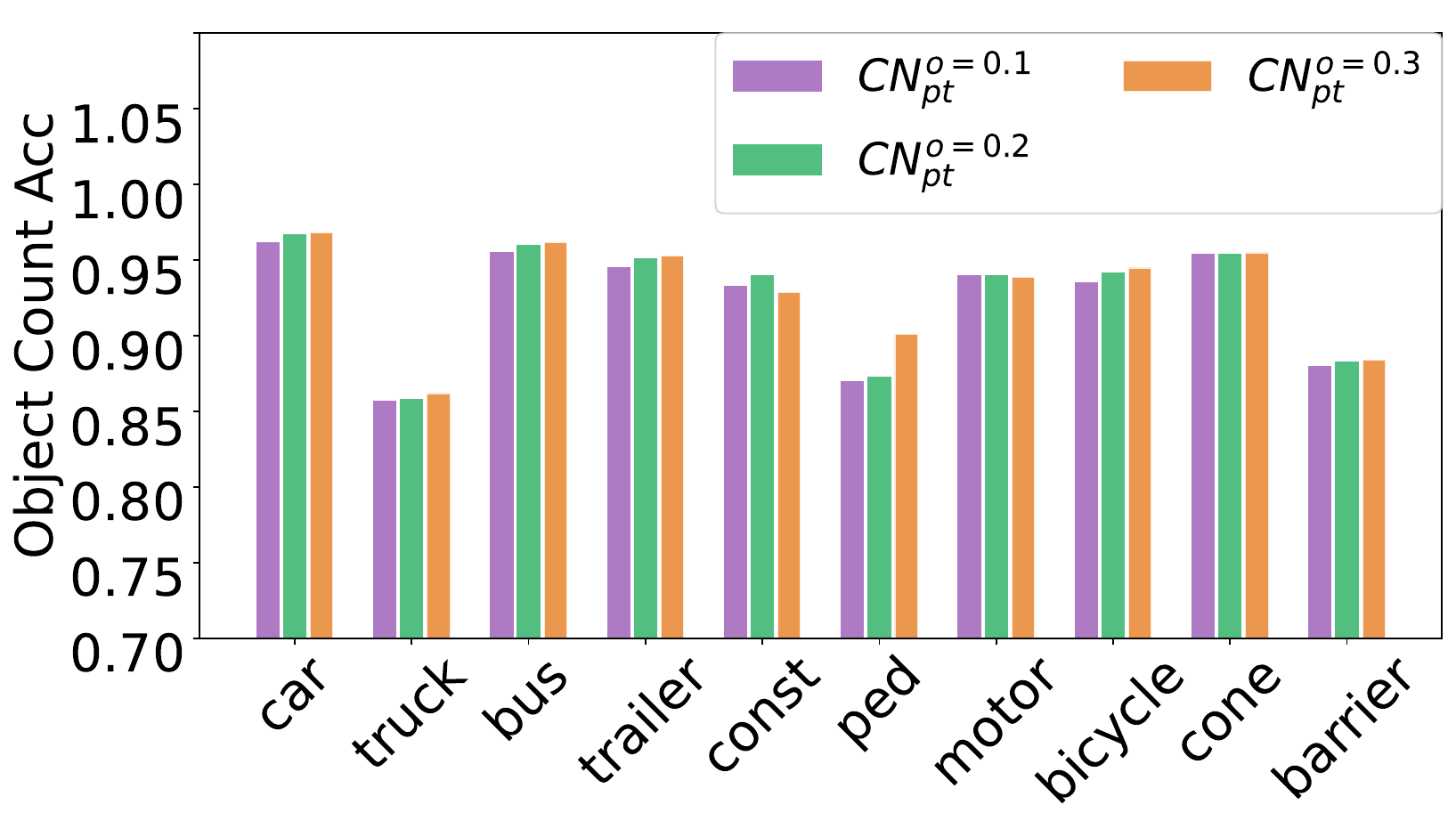}
    \caption{Different overlaps(nuScenes)}
    \label{fig: para_overlap}
\end{minipage}
\hfill 
\begin{minipage}{0.32\textwidth}
    \centering
    \includegraphics[width=\textwidth]{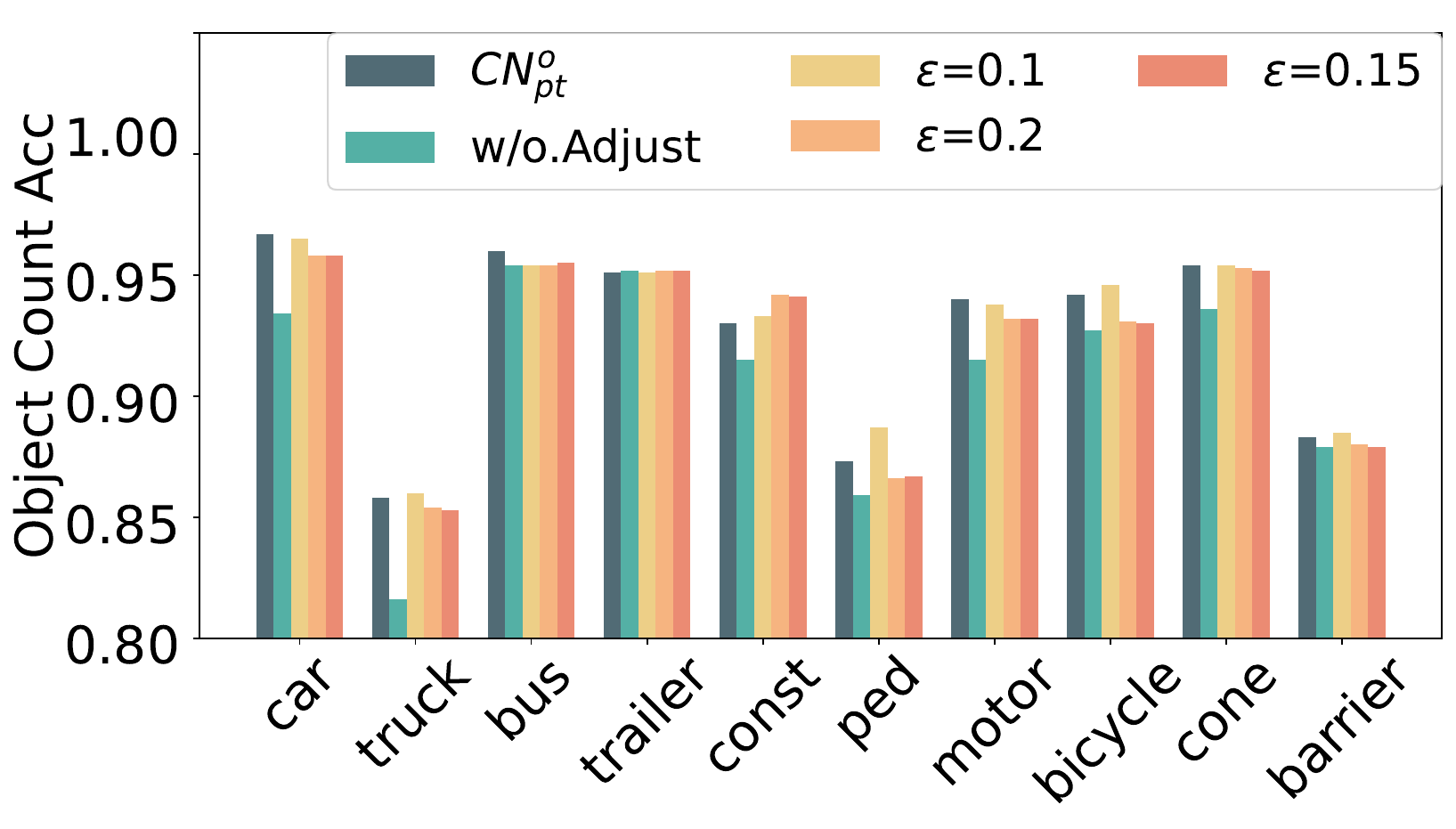}
    \captionof{figure}{Model selection(nuScenes)}
    \label{fig: model_selection}
\end{minipage}
\begin{minipage}{0.32\textwidth}
    \includegraphics[width=\textwidth]{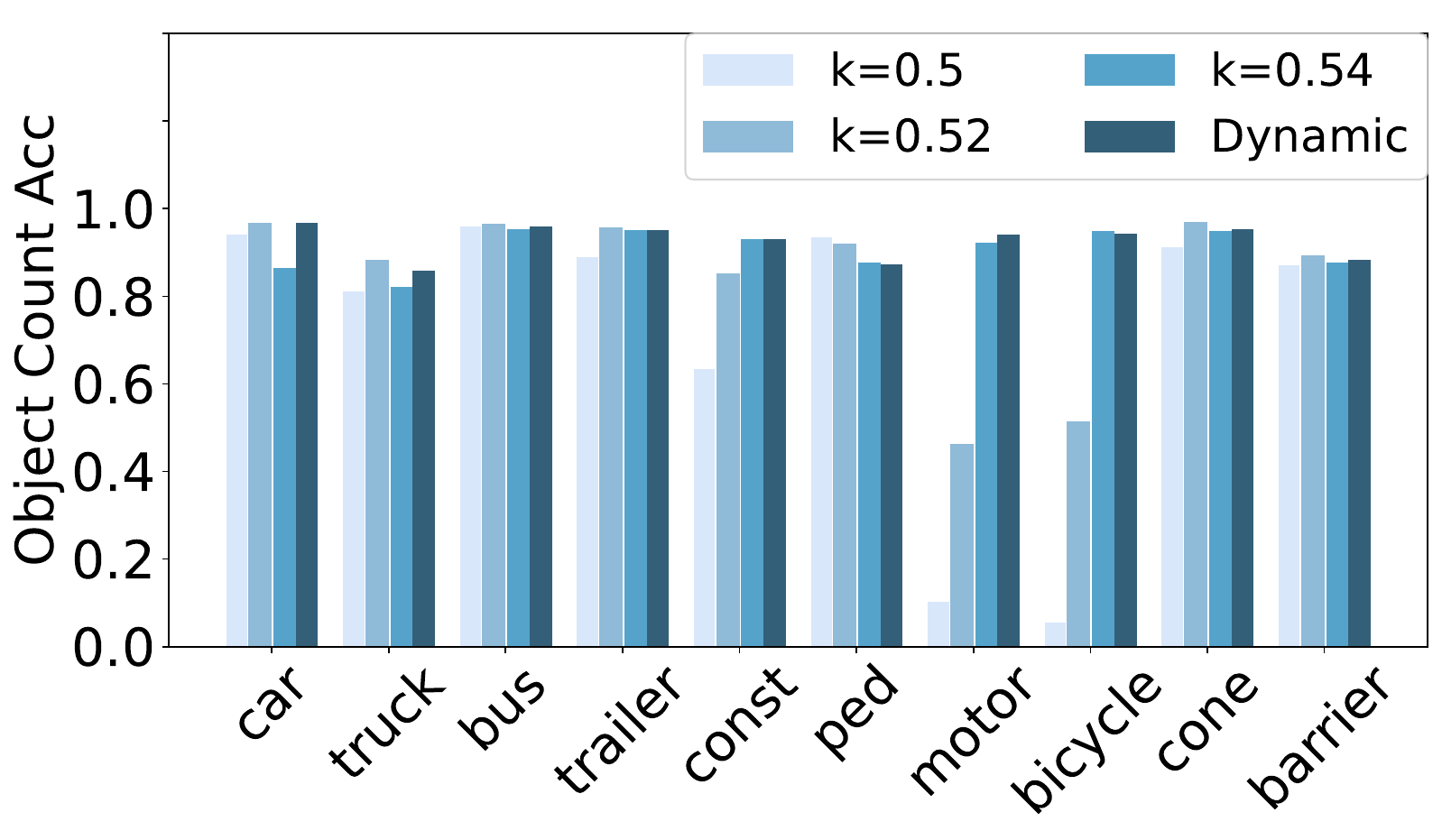}
    \centering
    \caption{Thresholding(nuScenes)}
    \label{fig: dynamic_threshold}
\end{minipage}
\end{figure*}

\textbf{(1) Partitioning without Overlap ($CN_{pt}$):} As shown in Figure~\ref{fig: para_partition}, increasing the number of partitions generally degrades performance for larger objects such as trucks and buses, likely due to their increased probability of being segmented. Conversely, smaller objects like pedestrians benefit from an increased number of partitions. 
Notably, performance declines sharply with nine partitions, indicating a negative effect on accuracy for most object categories. 

\noindent\textbf{(2) Partitioning with Overlap ($CN_{pt}^o$):} As shown in Figure~\ref{fig: para_partition}, the overlap can mitigate the negative impacts of partitioning. In most cases, with overlaps, the partitions enhance performance, particularly as the number of partitions increases. The benefits are obvious with 9 partitions compared with 2 partitions. 

\noindent\textbf{(3) Increasing Overlap Ratio:} As demonstrated in Figure~\ref{fig: para_overlap}, increasing the expansion ratio contributes little to the improvement of query accuracy. This finding indicates that while overlap helps, its benefits are not substantial through blind expansion.

Both Waymo and nuScenes provide 360$^{\circ}$ scenes, but Waymo covers a larger range (150.4m) compared to nuScenes (102.4m). In contrast, the KITTI dataset only captures a forward-facing view, with a range of 70.4m × 80m. Considering the differences in dataset scale, we analyze the impact of partitioning on query results across these datasets. As shown in Figure~\ref{fig: diff_partition_kitti} and Figure~\ref{fig: diff_partition_waymo}, Waymo, with its larger scene scale, tends to benefit more from partitioning and overlaps. In contrast, KITTI, which has a smaller scene scale, experiences a performance drop when using 9 partitions, which cannot be mitigated by overlap.

\subsection{Efficiency Study}

Table~\ref{table: time} presents per-frame processing times (in seconds) across three datasets: nuScenes, KITTI, and Waymo, for different model variants. Overall, processing time increases gradually from the baseline ($CN$) to more complex variants ($CN_{pt=2}$, $CN_{pt=4}$, $CN_{pt=9}$) and their optimized versions ($CN_{pt=2}^o$, $CN_{pt=4}^o$, $CN_{pt=9}^o$). Furthermore, on the nuScenes dataset, our method achieves comparable processing time to the baselines: 201 ms for TF and 154 ms for VN. While our solution yields similar results in terms of inference time, it demonstrates the advantage in effectiveness as discussed above.
\begin{table}[t]
\small
\begin{tabular}{|ll|ccc|}
\hline
                                         &       & \multicolumn{3}{c|}{COUNT} \\ \hline
                                         &       & nuScenes & KITTI  & Waymo  \\ \hline
\multicolumn{1}{|l|}{\multirow{4}{*}{\rotatebox{90}{baselines}}} & VN    & 0.658   & --     & 0.763 \\
\multicolumn{1}{|l|}{}                   & TF    & 0.683    & --     & --     \\
\multicolumn{1}{|l|}{}                   & SN    & 0.718    & --     & --     \\
\multicolumn{1}{|l|}{}                   & CP    & --       & 0.771 & --     \\ \hline
\multicolumn{1}{|l|}{\multirow{3}{*}{\rotatebox{90}{ours}}} & $CN$    & 0.759    & 0.7959 & 0.779 \\
\multicolumn{1}{|l|}{}                   & $CN_{pt}$  & 0.791    & 0.780 & 0.771 \\
\multicolumn{1}{|l|}{}                   & $CN_{pt}^o$ & \textbf{0.857}    & \textbf{0.846} & \textbf{0.807} \\ \hline
\end{tabular}

\caption{Overall query result of COUNT}
\vspace{-3em}

\label{table: all_count}
\end{table}

\subsection{Study of Model Selection}
\begin{table}[b!]
\small
\begin{tabular}{|p{0.8cm}p{0.3cm}p{0.7cm}p{0.7cm}p{0.7cm}p{0.7cm}p{0.7cm}l|}
\hline
         & $CN$    & $CN_{pt=2}$ & $CN_{pt=2}^o$ & $CN_{pt=4}$ & $CN_{pt=4}^o$ & $CN_{pt=9}$ & $CN_{pt=9}^o$ \\ \hline
nuScenes & 152 & 155 & 160  & 165 & 172  & 184 & 189  \\
KITTI    & 100 & 101 & 104  & 103 & 105  & 104 & 107  \\
Waymo    & 112 & 113 & 115  & 114 & 121  & 114 & 128  \\ \hline
\end{tabular}
\caption{Processing time per frame in milliseconds}
\label{table: time}
\end{table}

\begin{table}[t]
\centering
\small
\begin{tabular}{|ll|c|cc|cc|cc|}
\hline
\multicolumn{2}{|c|}{Model}& \multicolumn{1}{|r|}{w/o.Adjust} & \multicolumn{2}{c|}{$\epsilon$ = 0.1} & \multicolumn{2}{c|}{$\epsilon$ = 0.15} & \multicolumn{2}{c|}{$\epsilon$ = 0.2} \\ \hline
\multicolumn{1}{|l}{pt}& \multicolumn{1}{l|}{o}  & dist  & rate                     & dist                    & rate                      & dist                    & rate                     & dist                    \\ \hline
2 & 0                          & 1116  & 0.835                    & 339                     & 0.667                     & 168                     & 0.486                    & 90                      \\
2 & 0.1                     & 21    & 0.780                    & 279                     & 0.572                     & 274                     & 0.371                    & 343                     \\
2 & 0.2                    & 48    & 0.811                    & 43                      & 0.624                     & 35                      & 0.433                    & 32                      \\
4 & 0                         & 9     & 0.543                    & 556                     & 0.379                     & 743                     & 0.087                    & 752                     \\
4 & 0.1                     & 14    & 0.650                    & 9                       & 0.379                     & 7                       & 0.178                    & 8                       \\
4 & 0.2                     & 4739  & 0.617                    & 4736                    & 0.337                     & 4734                    & 0.145                    & 4735                    \\
9 & 0                          & 20    & 0.642                    & 0                       & 0.426                     & 3                       & 0.219                    & 3                       \\
9 & 0.1                   & 34    & 0.693                    & 48                      & 0.369                     & 53                      & 0.170                    & 54                      \\
9 & 0.2                     & 16    & 0.684                    & 7                       & 0.439                     & 35                      & 0.231                    & 0                       \\ \hline
\end{tabular}
\caption{Data distribution (abbr. dist) of model selection with adjustment with 9 models (nuScenes) }
\vspace{-3em}
\label{table: nus_adjustment}
\end{table}

Table~\ref{table: nus_adjustment} shows model selection results based on data distribution (i.e., frame allocation) with $\epsilon$ values from 0.1 to 0.2, where higher $\epsilon$ indicates a tighter bound. Key observations: (1) Model $CN_{pt=4}^{o=0.2}$ dominates, but frame allocation varies with adjustment. Without adjustment, many frames (e.g., 1116) are inefficiently allocated to suboptimal model $CN_{pt=2}^{o=0}$, while adjustment reallocates frames to more reliable models like $CN_{pt=2}^{o=0.1}$ (better for larger objects like buses), $CN_{pt=4}^{o=0}$, and $CN_{pt=9}^{o=0.1}$ (better for smaller objects like pedestrians). (2) Lower $\epsilon$ increases adjustment rates and true center deviations, while larger $\epsilon$ reduce adjustment influence; a balance is crucial. The selection of $\epsilon$ can be found in Appendix~\ref{app: model_selection}. (3) Evaluation via object count accuracy (Figure~\ref{fig: model_selection}) using $CN_{pt=4}^{o=0.2}$ as the baseline shows that, without model selection, performance worsens due to $CN_{pt=2}^{o=0}$. Adjustments improve balance across categories, trading $CN_{pt=4}^{o=0.2}$'s exceptional detection of cars and buses for gains in detecting pedestrians, bicycles, and barriers. For additional experiments on model selection, see Appendix~\ref{app: model_selection}.

\section{Conclusion}

In this paper, we present CounterNet, a method to facilitate querying 3D point cloud data with a focus on accurate object counting. By emphasizing object appearance detection and introducing a partition-based strategy with overlapping regions, we improve performance across varying object densities. Additionally, our dynamic model selection ensures robustness across scenarios. In the future, this research can be extended in directions: (1) Investigating more efficient solutions for querying point cloud data, optimizing both performance and scalability. (2) Leveraging the query tool to support data acquisition research, enabling the extraction of relevant data from a data pool to enhance model training for specific tasks.
\newpage

\noindent\textbf{ACKNOWLEDGEMENT}\\
Xiaoyu Zhang is a recipient of CSIRO Scholarship Stipend and Top-Up. 
This research was also supported partially by the Australian Government through the Australian Research Council's Discovery Projects funding scheme (projects DP240101211 and DP220101823).

\bibliographystyle{ACM-Reference-Format}
\bibliography{main}

\newpage
\appendix
\section{Table of Notations}

\begin{table}[h]
\small
\begin{tabular}{|ll|}

\hline
$\mathcal{P}$ & set of point cloud frames     \\
$p$                            & point cloud frame                 \\
$q$                            & query object                  \\
$op$                           & operations set                \\
$ct$                           & object count                  \\
$Y$                            & GT label for heatmap          \\
$\hat{Y}$                        & predicts for heatmap          \\
$(x, y)$                     & point coordinate              \\
$c$                            & object category               \\
$y$                            & GT object count               \\
$\hat{y}$                        & predict object count          \\
$k$                            & dynamic threshold for heatmap \\
$t$                            & fixed threshold for heatmap   \\
$r$                            & feature map region            \\
$pt$                           & partition number              \\
$o$                            & expanding overlapping ratio   \\
$M$                            & set of models                 \\
$m$                            & model                         \\
$\hat{\omega}$    & estimated center              \\
$\mu$          & true center                   \\
$\epsilon$      & expected deviation            \\
$\xi$          & sample variance               \\ \hline
\end{tabular}
\caption{Table of notations}
\label{table: notation}
\end{table}
\section{Motivation Example for the Study}
\label{app: motivation_example}

Our study is driven by the need to develop a query tool for point cloud data to address specific tasks. We use a real-world example to further illustrate the importance and necessity of this query tool, reinforcing the motivation behind our research.

Querying specific frames in point cloud data is crucial for facilitating target tasks. For example, to enhance a model’s ability to detect pedestrians, a direct and effective approach is to train or fine-tune the model using frames containing pedestrians rather than relying on the entire dataset. This targeted approach not only improves performance but also reduces computation costs. We conduct a simple experiment using the KITTI dataset. As shown in Figure~\ref{table: motivation}, we compare pedestrian detection performance under three conditions: (1) Training on the full dataset (3712 frames).
(2) Training on a randomly sampled subset of 1,000 frames. (3) Training on a carefully selected subset of frames containing pedestrians (about 955 frames).

The results demonstrate that using the carefully selected subset improved pedestrian detection performance, achieving similar or better results compared to the full dataset while reducing training costs by approximately two-thirds.

This experiment highlights the necessity of querying point cloud data, proving its potential to enhance task-specific performance and optimize resource utilization. 

\begin{table}[h]
\small
\centering
\begin{tabular}{|l|p{0.3cm}p{0.3cm}p{0.45cm}|p{0.35cm}p{0.35cm}p{0.5cm}|p{0.3cm}p{0.3cm}p{0.45cm}|}
\hline
                                                                  & \multicolumn{3}{c|}{Car} & \multicolumn{3}{c|}{Pedestrian}                  & \multicolumn{3}{c|}{Cyclist} \\ \cline{2-10} 
                                                                  & Easy   & Mod    & Hard   & Easy           & Mod            & Hard           & Easy     & Mod     & Hard    \\ \hline
\begin{tabular}[c]{@{}l@{}}Full \\ (3712 frames)\end{tabular}     & 69.42  & 61.44  & 58.33  & 33.88          & 31.96          & 31.25          & 63.47    & 48.28   & 46.11   \\
\begin{tabular}[c]{@{}l@{}}Random \\ (1000 frames)\end{tabular}   & 70.72  & 60.31  & 55.56  & 35.38          & 33.11          & 31.91          & 60.04    & 44.26   & 42.57   \\
\begin{tabular}[c]{@{}l@{}}Pedestrian\\ (955 frames)\end{tabular} & 70.38  & 58.38  & 54.63  & \textbf{37.88} & \textbf{36.11} & \textbf{34.53} & 59.46    & 43.81   & 42.49   \\ \hline
\end{tabular}
\caption{Detection results of training with different set}
\label{table: motivation}
\end{table}


\section{Data Model Design and Explanation for Query Type Selection}
\subsection{Data Model}
\label{app: data_model}
As illustrated below, our data model is designed with the NoSQL format, which brings more flexibility in aggregating data from different sources, such as different vehicles, making it more suitable for handling the diverse and complex nature of autonomous driving data. Object counting, being one of the most essential pieces of information required for querying~\cite{kang2022tasti,kang13blazeit,cao2022figo,bang2023seiden}, shapes the design of our data model, as shown below. We store key information from each frame, including object type, count, and the approximate positions of objects in a bird's eye view, ensuring that the data model is optimized for efficient querying.

\begin{lstlisting}[frame=single, breaklines, keepspaces, basicstyle=\footnotesize]
{ "frame_id": "frame123", 
  "timestamp": "2024-04-07 15:43:02.8924030000",
  "vehicle_id": "vehicle_00000000",
  "objects": [ 
	{ "type": "car", 
	   "count": "10", 
	"position":[{"x": 1.5, "y": 2.1}, ...]},
    ...] 
}

\end{lstlisting}

\section{Methodology}

\subsection{Steps of Feature Map Partition with Overlap}
\label{app: overlap_partition}

\begin{algorithm}
\small
    \caption{Inference with Overlap Partition}
    \label{alg: overlap}
    \KwIn{feature map $F$, expanding ratio $\delta$, radius $\gamma$}
    \KwOut{centers}
    $R = getPartitions(F) $\; \label{ol: partition}
    \tcp{Expand regions}
    \ForEach{$r \in R$}{
    $expand\_w = r.width \times \delta$ \; \label{ol: expand1}
    $expand\_h = r.height \times \delta$ \;\label{ol: expand2}
    $r.x_{start} = max(0, r.x_{start}-expand\_w)$ \;\label{ol: expand3}
    $r.y_{start} = max(0, r.y_{start}-expand\_h)$ \;
    $r.x_{end} = min(F.width, r.x_{end}+expand\_w)$ \;
    $r.y_{end} = min(F.height, r.y_{end}+expand\_h)$ \; \label{ol: expand4}
    }
    $H=getHeatmaps(R)$ \label{ol: hm} \;\label{ol: heatmap}
    $map = zeros((F.width, F.height))$\tcp{init map of 0} \label{ol: center1}
    \ForEach{$h \in H$}{
    C = getCentersFromHeatmap(h) \;
    \tcp{region coord -> map coord}
    C = transCoordinate(C) \;
    \ForEach{$c \in centers$}{
        map[c.x, c.y] = 1
    } \label{ol: center2}
    }
    \tcp{Combine duplicate centers}
    $centers = getCenters(map)$\; \label{ol: combine1}
    $new\_centers = \emptyset$ \;
    \While{centers}
    {
    $current\_center = centers.pop(0)$\;
    $new\_centers.add(current_c)$\;
    $centers = combineCenters(centers, r)$\; \label{ol: combine2}
    }
    return $new\_centers$
\end{algorithm}

The training process and loss calculation follow the same procedure as CounterNet with Partition introduced in Section~\ref{sec: counternet_partition}. However, the inference process involves additional steps to account for overlapping partitions, as shown in Figure~\ref{fig: para_overlap} and Algorithm \ref{alg: overlap}: (1) Generate Expanded Partitioned Heatmaps. Given a feature map, we first divide it into partitions (Line \ref{ol: partition}). Based on the expansion ratio $\delta$, we calculate the required width and height for the expansion (Lines \ref{ol: expand1}-\ref{ol: expand2}) and determine the region coordinates with the expanded width and height (Lines \ref{ol: expand3}-\ref{ol: expand4}). (2) Extract Centers from Partitions and Merge. We generate the heatmap for each partitioned region (Line \ref{ol: heatmap}). Afterward, the centers extracted from each heatmap are merged into a single map (Lines \ref{ol: center1}-\ref{ol: center2}). (3) Remove Duplicate Center Points. Using a predefined radius $\gamma$, we merge centers that fall within this radius (Lines \ref{ol: combine1}-\ref{ol: combine2}).

\subsection{Steps of Model Selection}
\label{app: model_selection}
The overall process of model selection is explained in Algorithm~\ref{alg: model_selection}, which contains three main steps: (1) Model center representation. (2) Center adjustment with
confidence. (3) Find the optimal model.

\begin{algorithm}[t!]
\small
\SetAlgoLined
\KwIn{model set $M$, threshold $\epsilon$, training point cloud set $\mathcal{P}_{train}$, eval point cloud set $\mathcal{P}_{eval}$}

\tcc{Step 1: Model Center Representation}
\tcc{Initialize assignment sets}
Initialize $m.A \gets \emptyset$ for each $m \in M$ 

\ForEach{$p \in \mathcal{P}$}{
    $best\_model \gets None$ \;
    $best\_acc \gets -\infty$ \;

    \ForEach{$m \in M$}{
        $acc \gets \text{EvaluateModel}(m, p)$ \tcp*{Evaluate model $m$ on $p$}

        \If{$acc > best\_acc$}{
            $best\_acc \gets acc$ \;
            $best\_model \gets m$ \;
        }
    }

    $best\_model.A \gets best\_model.A \cup \{p\}$

}

\tcc{Step 2: Center Adjustment with Confidence}
Initialize $m.\hat{\omega} \gets \emptyset$ and $m.P \gets \emptyset$ \;

\ForEach{$m \in M$}{
    \tcc{Number of assigned point clouds}
    $n \gets |m.A|$ 
    
    
    \tcc{estimated center}
    $m.\hat{\omega} \gets \frac{1}{n} \sum\limits_{p \in m.A} p.f$ 

    \tcc{sample variance}
    $\xi^2 \gets \frac{1}{n} \sum\limits_{p \in m.A} (p.f - m.\hat{\omega})^2$ 

    \tcc{confidence probability using Chernoff bound}
    $m.c \gets \exp\left(-\frac{n \epsilon^2}{2 \xi^2}\right)$ 
}

\tcc{Step 3: Find the optimal model}
\ForEach{$p \in \mathcal{P}_{val}$}{
    \tcc{Track best model and minimum adjusted distance}
    Initialize $m^* \gets None$, $d_{min} \gets \infty$ 
    
    \ForEach{$m \in M$}{
    \tcc{Euclidean distance}
    $d \gets \text{ComputeDistance}(p.f, m.\hat{\omega})$ 

    \tcc{Adjust distance}
    $d_{adjusted} \gets d \times m.c$ 

    \If{$d_{adjusted} < d_{min}$}{
        $d_{min} \gets d_{adjusted}$; $m^* \gets m$ 
    }
    }
    $m^*(p)$

}

\caption{Model Selection}
\label{alg: model_selection}
\end{algorithm}







\section{Additional Experiments on nuScenes}
\label{app: nuscenes}
The nuScenes dataset includes 28,130 samples in the training set and 6,019 samples in the validation set. We train and evaluate the model with 10 categories of objects, including car, truck, construction\_vehicle (const), bus, trailer, barrier, motorcycle (moto), bicycle (bicyc), pedestrian (ped), and traffic\_cone (cone).
\begin{table}[t]
\centering
\small
\begin{tabular}{p{1.4cm}p{0.4cm}p{0.5cm}p{0.62cm}p{0.62cm}p{0.62cm}p{0.62cm}p{0.62cm}p{0.3cm}}
\hline
\# of objects                      & 1-5  & 6-10 & 11-15 & 16-20 & 21-30 & 31-40 & 41-50 & $\geq$51 \\ \hline
car                   & 1953 & 1794 & 1070  & 578   & 223   & 69    & 19    & 8            \\
pedestrian            & 2863 & 724  & 231   & 197   & 113   & 30    & 6     & 12           \\
barrier               & 463  & 329  & 280   & 190   & 113   & 33    & 16    & 4            \\
traffic\_cone         & 1408 & 417  & 104   & 48    & 71    & 10    & 1     & --           \\
truck                 & 3225 & 248  & 63    & 6     & --    & --    & --    & --           \\
bus                   & 1504 & 12   & --    & --    & --    & --    & --    & --           \\
trailer               & 818  & 81   & 37    & 6     & --    & --    & --    & --           \\
const\_vehicle & 1134 & --   & --    & --    & --    & --    & --    & --           \\
motorcycle            & 1192 & 40   & --    & --    & --    & --    & --    & --           \\ \hline
\end{tabular}
\caption{Statistic of nuScenes dataset}
\label{table: nus_distribution}
\end{table}
\subsection{Data Distribution of Dataset}
\label{app: nus_distribution}

Table~\ref{table: nus_distribution} demonstrates the data distribution of the nuScenes dataset. In this table, we present the statistics showing the number of frames that contain a specific number of objects (as indicated in the first row).

\subsection{Categorical Query Results}
By analyzing the results from RETRIEVAL-SINGLE (Table~\ref{table: result_select}), COUNT and AGGREGATION (Table~\ref{table: result_count_agg}), we observe: (1) CounterNet (CN) Performance: CounterNet outperforms detection models, demonstrating its outstanding performance in extracting query-specific information. (2) Impact of Partitioning: Partitioning improves accuracy for small-object categories like cars and barriers, while performance remains stable for pedestrians. However, it reduces accuracy for larger objects (e.g., trucks, buses) due to increased segmentation across partitions, aligning with our hypothesis that partitioning benefits small or numerous objects but hinders performance for infrequent, large objects. (3) Impact of Overlapping: Overlapped partitions improve performance across all categories, mitigating partitioning drawbacks (e.g., segmentation issues) while maintaining their advantages for counting small or numerous objects.
\begin{table*}[t!]
\centering
\small
\begin{tabular}{|p{0.5cm}cccccc|cccccc|ccc|}
\hline
        & \multicolumn{6}{c|}{RETRIEVAL-SINGLE(10\% tolerance)} & \multicolumn{6}{c|}{RETRIEVAL-SINGLE(5\% tolerance)}       \\ \hline
        & VN       & TF    & SN   & $CN$      & $CN_{pt}$   & $CN_{pt}^{o}$   & VN      & TF   & SN   & $CN$      & $CN_{pt}$   & $CN_{pt}^{o}$   \\ \hline
car     & 0.801    & 0.615  & 0.861 & 0.807   & 0.919    & 0.967     & 0.536   & 0.506   &0.635& 0.601   & 0.774    & 0.861        \\
truck   & 0.778    & 0.621  & 0.805 & 0.802   & 0.799    & 0.858     & 0.594   & 0.610   &0.668& 0.659   & 0.609$^\downarrow$    & 0.705        \\
bus     & 0.917    & 0.959  & 0.982 & 0.957   & 0.944    & 0.960     & 0.917   & 0.959   &0.982& 0.957   & 0.944$^\downarrow$    & 0.960          \\
trailer & 0.915    & 0.925  & 0.937 & 0.953   & 0.950    & 0.951     & 0.815   & 0.909   &0.890& 0.912   & 0.880$^\downarrow$    & 0.871        \\
const   & 0.845    & 0.886  & 0.924 & 0.929   & 0.910    & 0.931     & 0.849   & 0.886   &0.924& 0.929   & 0.909$^\downarrow$    & 0.810        \\
ped     & 0.711    & 0.640  & 0.773 & 0.870   & 0.869    & 0.873     & 0.469   & 0.553   &0.579& 0.741   & 0.735$^\downarrow$    & 0.748        \\
moto    & 0.728    & 0.867  & 0.906 & 0.933   & 0.886$^\downarrow$    & 0.940     & 0.728   & 0.866   &0.906& 0.933   & 0.886$^\downarrow$    & 0.940        \\
bicyc   & 0.632    & 0.828  & 0.893 & 0.940   & 0.898$^\downarrow$    & 0.942     & 0.632   & 0.907   &0.893& 0.940   & 0.898$^\downarrow$    & 0.942        \\
cone    & 0.706    & 0.722  & 0.824 & 0.933   & 0.926$^\downarrow$    & 0.954     & 0.528   & 0.717   &0.727& 0.875   & 0.874$^\downarrow$    & 0.867        \\
barrier & 0.761    & 0.802  & 0.810 & 0.874   & 0.883    & 0.883     & 0.640   & 0.755   &0.738& 0.811   & 0.823    & 0.824        \\ \hline
\end{tabular}
\caption*{\footnotesize $\downarrow$ indicates a performance drop compared to $CN$.}
 \vspace{-0.4cm}
\captionof{table}{Result of RETRIEVAL-SINGLE (nuScenes)}
\label{table: result_select}

\end{table*}
\begin{table*}
\small
\begin{tabular}{|p{0.45cm}p{0.45cm}p{0.45cm}p{0.45cm}ccc|p{0.45cm}p{0.45cm}p{0.45cm}ccc|p{0.45cm}p{0.45cm}p{0.45cm}ccc|}
\hline
        & \multicolumn{6}{c|}{COUNT}               & \multicolumn{6}{c|}{AGG (absolute)} & \multicolumn{6}{c|}{AGG (q-error)}                 \\ \hline
        & VN    & TF    & SN& $CN$    & $CN_{pt}$ & $CN_{pt}^o$ & VN    & TF   & SN & $CN$    & $CN_{pt}$ & $CN_{pt}^o$ & VN    & TF  &SN  & $CN$    & $CN_{pt}$ & $CN_{pt}^o$ \\ \hline
car     & 0.552 & 0.355 & 0.695&0.947 & 0.927$^\downarrow$  & 0.953   & 4.021 & 6.488 &3.213& 3.986 & 2.357  & \textbf{0.662} &1.478 & 1.736 &1.634& 1.401 & 1.403$^\uparrow$ & \textbf{1.290}  \\
truck   & 0.309 & 0.431 & 0.602&0.351 & 0.588  & 0.592   & 1.544 & 2.568 &1.385& 1.157 & 0.723  & \textbf{0.506}   & 1.979 & 2.222 &1.983& 1.828 & 1.842$^\uparrow$ & \textbf{1.522}\\
bus     & 0.950 & 0.994 & 0.995&0.859 & 0.857$^\downarrow$  & 0.860    & 0.401 & 0.220 &0.112& 0.228 & 0.259  & \textbf{0.183}  & 1.000 & 1.000 &1.292& 1.113 & 1.065 & \textbf{1.000} \\
trailer & 0.562 & 0.858 & 0.798&0.787 & 0.829  & 0.865   & 0.786 & 0.703 &0.588& 0.373 & \textbf{0.355}  & 0.390  & \textbf{1.001} & 1.256 &1.075& 1.057 & 1.144$^\uparrow$ & 1.240 \\
const   & 0.920 & 0.947 & 0.904&0.890 & 0.851$^\downarrow$  & 0.936   & 0.628 & 0.475 &0.287& \textbf{0.221} & 0.235$^\uparrow$  & 0.275$^\uparrow$   & 1.220 & 1.180 &1.166& 1.112 & 1.131 & \textbf{1.087}\\
ped     & 0.262 & 0.307 & 0.425&0.671 & 0.732  & 0.745   & 6.039 & 7.537 &4.726& 2.768 & 2.636  & \textbf{2.516} & 2.560 & 2.913 &2.023& 2.157 & 2.143 & \textbf{1.967}  \\
moto    & 0.866 & 0.920 & 0.887&0.825 & 0.830  & 0.884   & 1.021 & 0.573 &0.406& \textbf{0.190} & 0.304$^\uparrow$  & 0.201 & 1.572 & 1.232 &1.272& 1.315 & 1.293 & \textbf{1.061}  \\
bicyc   & 0.771 & 0.946 & 0.956&0.949 & 0.948$^\downarrow$  & 0.950   & 1.315 & 0.649 &0.437& 0.144 & 0.330$^\uparrow$  & \textbf{0.140} & 1.764 & 1.482 &1.201& 1.167 & 1.177$^\uparrow$ & \textbf{1.064}  \\
cone    & 0.180 & 0.251 & 0.433&0.572 & 0.586  & 0.892   & 5.189 & 4.854 &3.055& 1.091 & 1.123$^\uparrow$  & \textbf{0.898} & 3.161 & 2.725 &2.288& \textbf{1.550} & 1.598$^\uparrow$ & 1.944$^\uparrow$  \\
barrier & 0.208 & 0.410 & 0.494&0.745 & 0.765  & 0.896   & 5.429 & 4.029 &3.574& 2.128 & 2.115  & \textbf{1.670} & 2.303 & 1.996 &1.621& 1.464 & 1.490$^\uparrow$ & \textbf{1.041}  \\ \hline
\end{tabular}
\centering

\caption*{\footnotesize $\downarrow$ indicates a performance drop for COUNT, while $\uparrow$ means a performance drop for AGG, compared to $CN$.}
\vspace{-0.4cm}
\captionof{table}{Result of COUNT and AGGREGATION (nuScenes)}
\label{table: result_count_agg}

\end{table*}

\begin{table*}[t!]
\small
\centering
\begin{tabular}{|lcccc|cccc|cccc|cccc|}
\hline
        & \multicolumn{4}{c|}{RETRIEVE-SINGLE(10\% tolerance)} & \multicolumn{4}{c|}{COUNT}                       & \multicolumn{4}{c|}{AGG(absolute)}  & \multicolumn{4}{c|}{AGG(q-error)}                \\ \hline
        & CP              & $CN$      & $CN_{pt}$  & $CN_{pt}^o$         & CP    & $CN$             & -$CN_{pt}$ & $CN_{pt}^o$         & CP             & $CN$             & $CN_{pt}$ & $CN_{pt}^o$ & CP   & $CN$             & $CN_{pt}$ & $CN_{pt}^o$\\ \hline
car     & \textbf{0.952}  & 0.9156  & 0.929   & 0.934          & 0.692 & \textbf{0.705} & 0.681  & 0.701          & \textbf{0.135} & 0.280          & 2.25   & 0.172  & 1.134 & 1.159 & 1.198 & 1.154\\
ped     & 0.602           & 0.675   & 0.639   & \textbf{0.792} & 0.776 & 0.778          & 0.781  & \textbf{0.828} & 0.704          & \textbf{0.167} & 3.07   & 0.447  & 1.724 & 1.082 & 1.004 & 1.0\\
cyclist & 0.760           & 0.797   & 0.773   & \textbf{0.812} & 0.901 & 0.925          & 0.922  & \textbf{0.931} & 0.152          & \textbf{0.121} & 2.183  & 0.211  & 1.0 & 1.0 & 1.0 & 1.0\\ \hline
\end{tabular}
\caption{Query result on KITTI dataset}
\label{table: kitti_query}
\end{table*}


\begin{table*}[t!]
\small
\centering
\begin{tabular}{|lcccc|cccc|cccc|cccc|}
\hline
\multicolumn{5}{|c|}{RETRIEVAL-SINGAL(10\% tolerance)} & \multicolumn{4}{c|}{COUNT}    & \multicolumn{4}{c|}{AGG(absolute)} & \multicolumn{4}{c|}{AGG(q-error)}\\ \hline
              & VN       & $CN$      & $CN_{pt}$   & $CN_{pt}^o$   & VN    & $CN$    & $CN_{pt}$ & $CN_{pt}^o$ & VN     & $CN$     & $CN_{pt}$  & $CN_{pt}^o$ & VN     & $CN$     & $CN_{pt}$  & $CN_{pt}^o$ \\ \hline
vehicle       & 0.777    & 0.817   & 0.818   & 0.861   & 0.828 & 0.814 & 0.799 & 0.835 & 8.927  & 6.385  & 3.595  & 3.420 & 1.269 & 1.242 & 1.199 & 1.196\\
pedestrian    & 0.776    & 0.814   & 0.824   & 0.852   & 0.724 & 0.736 & 0.738 & 0.799 & 8.888  & 6.501  & 3.523  & 3.342 & 1.271 & 1.244 & 1.193 & 1.197\\
cyclist       & 0.784    & 0.817   & 0.820   & 0.852   & 0.739 & 0.788 & 0.776 & 0.788 & 8.839  & 6.289  & 3.567  & 3.143 & 1.265 & 1.258 & 1.205 & 1.199\\ \hline
\end{tabular}
\caption{Query result on Waymo dataset}
\label{table: waymo_query}
\end{table*}
\begin{table*}[t!]
\centering
\small
\begin{tabular}{|lll|p{0.3cm}p{0.3cm}p{0.3cm}p{0.3cm}p{0.25cm}|p{0.3cm}p{0.3cm}p{0.3cm}p{0.3cm}p{0.25cm}|p{0.3cm}p{0.3cm}p{0.3cm}p{0.3cm}p{0.25cm}|}
\hline
\multicolumn{3}{|c|}{Query Information}                                                                       & \multicolumn{5}{c|}{ACC}                                                                                                           & \multicolumn{5}{c|}{Recall}                                                                                                        & \multicolumn{5}{c|}{Precision}                                                                                                     \\ \hline
 & \multicolumn{2}{l|}{Predict  \hspace*{1.7cm}  Selectivity}                                              & \multicolumn{1}{l}{VN} & \multicolumn{1}{l}{TF} & \multicolumn{1}{l}{$CN$} & \multicolumn{1}{l}{$CN_{pt}$} & \multicolumn{1}{l|}{$CN_{pt}^o$} & \multicolumn{1}{l}{VN} & \multicolumn{1}{l}{TF} & \multicolumn{1}{l}{$CN$} & \multicolumn{1}{l}{$CN_{pt}$} & \multicolumn{1}{l|}{$CN_{pt}^o$} & \multicolumn{1}{l}{VN} & \multicolumn{1}{l}{TF} & \multicolumn{1}{l}{$CN$} & \multicolumn{1}{l}{$CN_{pt}$} & \multicolumn{1}{l|}{$CN_{pt}^o$} \\
Q1    & RETRIEVAL(car\textgreater{}5)                                                                   & 67.9\% & 0.812                  & 0.768                  & 0.614                  & 0.778                     & \textbf{0.850}              & 0.815                  & 0.769                  & 0.996                  & 0.972                     & 0.929                       & 0.995                  & 0.997                  & 0.615                  & 0.796                     & 0.908                       \\
Q2    & RETRIEVAL(barrier\textgreater{}3)                                                               & 76.2\% & 0.639                  & 0.748                  & 0.859                  & \textbf{0.866}            & 0.764                       & 0.954                  & 0.953                  & 0.905                  & 0.893                     & 0.934                       & 0.641                  & 0.752                  & 0.943                  & 0.967                     & 0.807                       \\
Q3    & RETRIEVAL(bus\textgreater{}0)                                                                   & 33.2\% & 0.488                  & 0.507                  & \textbf{0.553}         & 0.445                     & 0.557                       & 0.504                  & 0.628                  & 0.898                  & 0.762                     & 0.759                       & 0.940                  & 0.949                  & 0.591                  & 0.515                     & 0.632                       \\
Q5    & \begin{tabular}[c]{@{}l@{}}RETRIEVAL(car\textgreater{}5,\\ ped\textgreater{}0)\end{tabular}       & 31.5\% & 0.635                  & 0.604                  & 0.520                  & 0.624                     & \textbf{0.702}              & 0.638                  & 0.605                  & 0.967                  & 0.857                     & 0.841                       & 0.962                  & 0.966                  & 0.529                  & 0.697                     & 0.809                       \\
Q6    & \begin{tabular}[c]{@{}l@{}}RETRIEVAL(truck\textgreater{}0,\\ barrier\textgreater{}5)\end{tabular} & 16.8\% & 0.404                  & 0.448                  & 0.332                  & 0.465                     & \textbf{0.529}              & 0.411                  & 0.452                  & 0.761                  & 0.882                     & 0.678                       & 0.964                  & 0.977                  & 0.371                  & 0.496                     & 0.706                       \\ \hline
\end{tabular}
\caption{Case study of different query scenario (nuScenes)}
\label{table: case_study}
\end{table*}

\subsection{Model Selection}
\label{app: model_selection}

\begin{figure*}
\begin{minipage}{0.33\textwidth}
    \includegraphics[scale=0.2]{images/adjustment.pdf}
    \centering
    \caption{Model selection with 9 models (nuScenes)}
    \label{fig: adjustment_9}
\end{minipage}
\hfill 
\begin{minipage}{0.32\textwidth}
    \includegraphics[scale=0.2]{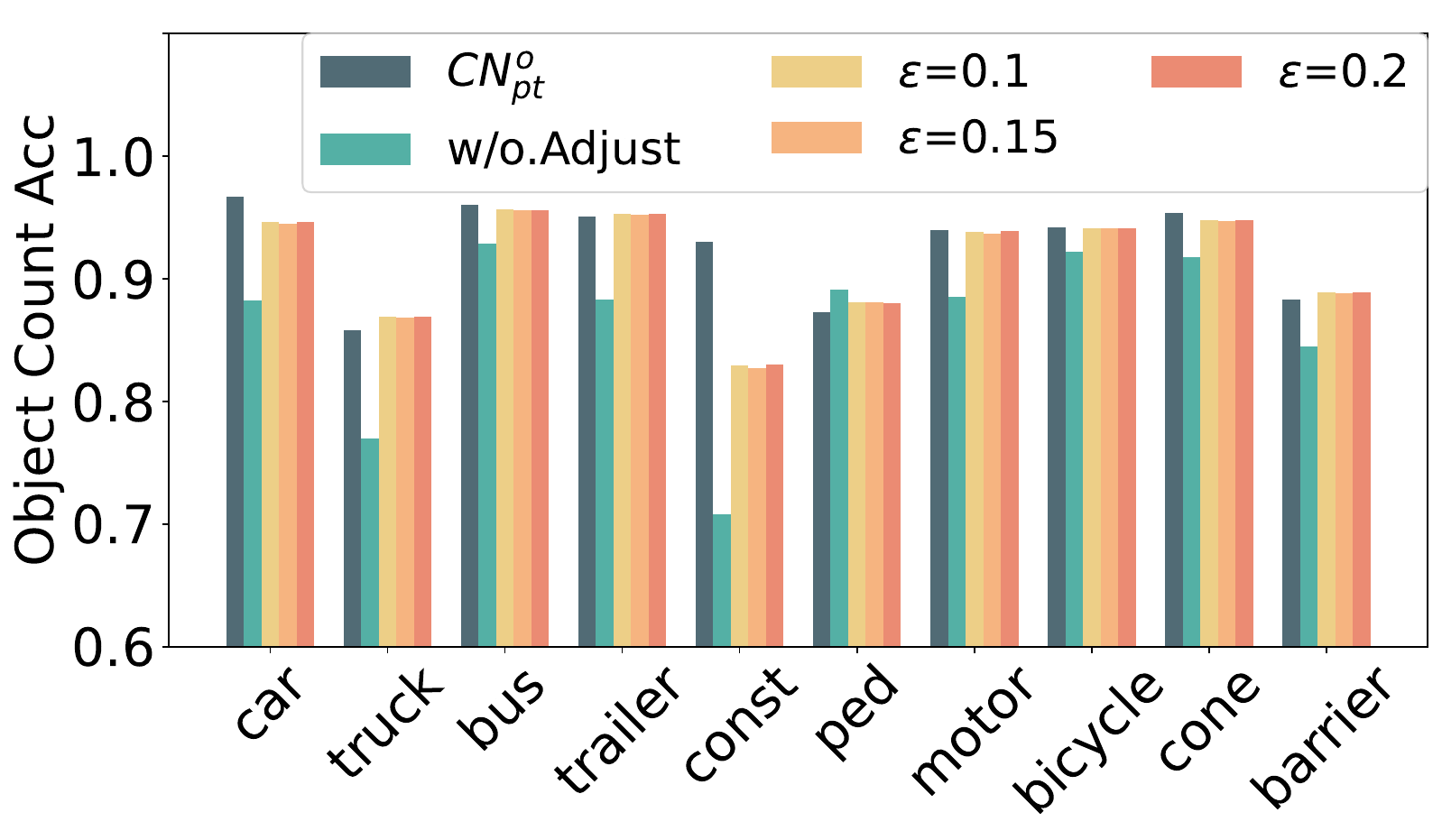}
    \centering
    \caption{Model selection with 6 models (nuScenes)}
    \label{fig: adjustment_6}
\end{minipage}
\begin{minipage}{0.32\textwidth}
    \includegraphics[scale=0.2]{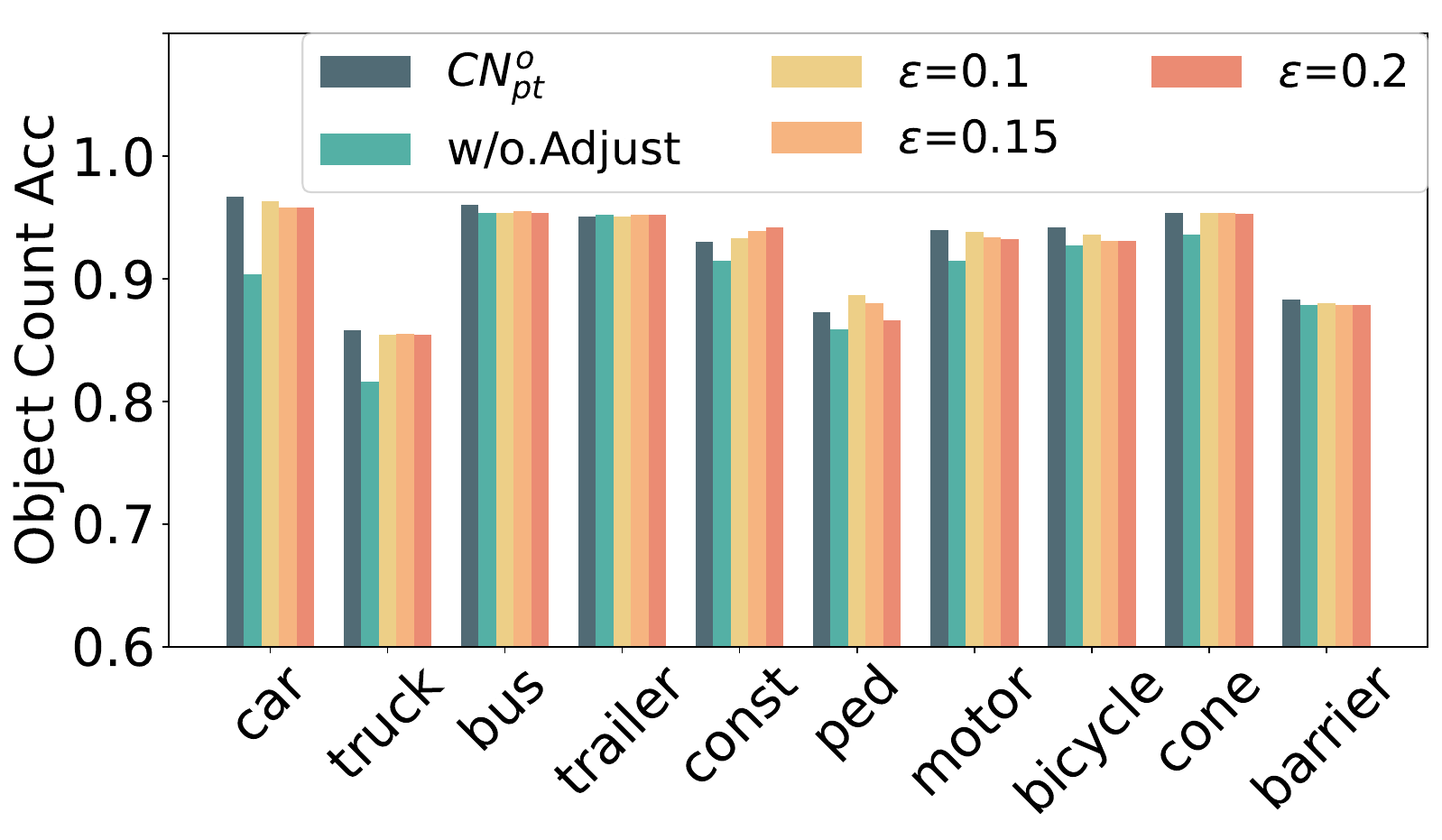}
    \centering
    \caption{Model selection with emphasis on car \& pedestrian (nuScenes)}
    \label{fig: adjustment_9_car_ped}
\end{minipage}
\end{figure*}

\begin{table}[t]
\centering
\small
\begin{tabular}{|cc|cc|cc|cc|}
\hline
\multicolumn{2}{|r|}{no adjustment} & \multicolumn{2}{c|}{$\epsilon$ = 0.1} & \multicolumn{2}{c|}{$\epsilon$ = 0.15} & \multicolumn{2}{c|}{$\epsilon$ = 0.2} \\ \hline
\multicolumn{1}{|l}{Model}  & dist  & rate                     & dist                    & rate                      & dist                    & rate                     & dist                    \\ \hline
p2                          & 1116  & 0.835                    & 339                     & 0.667                     & 168                     & 0.486                    & 90                      \\
p2-o0.1                     & 21    & 0.780                    & 279                     & 0.572                     & 274                     & 0.371                    & 343                     \\
p2-o0.2                     & 48    & 0.811                    & 43                      & 0.624                     & 35                      & 0.433                    & 32                      \\
p4                          & 9     & 0.543                    & 556                     & 0.379                     & 743                     & 0.087                    & 752                     \\
p4-o0.1                     & 14    & 0.650                    & 9                       & 0.379                     & 7                       & 0.178                    & 8                       \\
p4-o0.2                     & 4739  & 0.617                    & 4736                    & 0.337                     & 4734                    & 0.145                    & 4735                    \\
p9                          & 20    & 0.642                    & 0                       & 0.426                     & 3                       & 0.219                    & 3                       \\
p9-o0.1                     & 34    & 0.693                    & 48                      & 0.369                     & 53                      & 0.170                    & 54                      \\
p9-o0.2                     & 16    & 0.684                    & 7                       & 0.439                     & 35                      & 0.231                    & 0                       \\ \hline
\end{tabular}
\caption{Data distribution (abbr. dist) of model selection with adjustment with 9 models
(nuScenes)}
\label{table: adjustment_9}
\end{table}
\begin{table}[t]
\small
\centering
\begin{tabular}{|ll|c|cc|cc|cc|}
\hline
\multicolumn{2}{|c|}{Model}&\multicolumn{1}{|r|}{w/o.Adjust} & \multicolumn{2}{c|}{$\epsilon$ = 0.1} & \multicolumn{2}{c|}{$\epsilon$ = 0.15} & \multicolumn{2}{c|}{$\epsilon$ = 0.2} \\ \hline
\multicolumn{1}{|l}{pt}& \multicolumn{1}{l|}{o}  & dist  & rate                     & dist                    & rate                      & dist                    & rate                     & dist                    \\ \hline
2&0                          & 0     & 0.718                    & 0                       & 0.475                     & 0                       & 0.266                    & 0                       \\
2&0.1                     & 1171  & 0.631                    & 641                     & 0.355                     & 561                     & 0.159                    & 713                     \\
2&0.2                     & 60    & 0.633                    & 54                      & 0.358                     & 53                      & 0.161                    & 56                      \\
9&0                          & 6  & 0.616                    & 0                       & 0.336                     & 0                       & 0.144                    & 1                       \\
9&0.1                     & 4752     & 0.405                    & 5323                    & 0.131                     & 5404                    & 0.026                    & 5237                    \\
9&0.2                     & 30    & 0.497                    & 1                       & 0.207                     & 1                       & 0.061                    & 12                      \\ \hline
\end{tabular}
\caption{Data distribution (abbr. dist) of model selection with adjustment with 6 models (nuScenes)}
\label{table: adjustment_6}
\end{table}

\begin{table}[t]
\centering
\small
\begin{tabular}{|ll|c|cc|cc|cc|}
\hline
\multicolumn{2}{|c|}{Model} & \multicolumn{1}{|r|}{w/o.Adjust} &\multicolumn{2}{c|}{$\epsilon$ = 0.1} & \multicolumn{2}{c|}{$\epsilon$ = 0.15} & \multicolumn{2}{c|}{$\epsilon$ = 0.2} \\ \hline
\multicolumn{1}{|l}{pt}  & \multicolumn{1}{l|}{o}  & dist  & rate                     & dist                    & rate                      & dist                    & rate                     & dist                    \\ \hline
2&0                          & 19    & 0.833                    & 14                      & 0.663                     & 12                      & 0.481                    & 9                       \\
2&0.1                     & 1133  & 0.809                    & 656                     & 0.621                     & 461                     & 0.429                    & 414                     \\
2&0.2                     & 20    & 0.837                    & 18                      & 0.670                     & 17                      & 0.490                    & 17                      \\
4&0                          & 23    & 0.660                    & 22                      & 0.393                     & 24                      & 0.190                    & 59                      \\
4&0.1                     & 22    & 0.634                    & 17                      & 0.358                     & 13                      & 0.161                    & 14                      \\
4&0.2                     & 4728  & 0.578                    & 4734                    & 0.292                     & 4735                    & 0.112                    & 4734                    \\
9&0                          & 26    & 0.628                    & 19                      & 0.352                     & 18                      & 0.156                    & 19                      \\
9&0.1                     & 13    & 0.593                    & 502                     & 0.309                     & 702                     & 0.124                    & 717                     \\
9&0.2                     & 33    & 0.691                    & 35                      & 0.435                     & 35                      & 0.228                    & 34                      \\ \hline
\end{tabular}
\caption{Data distribution (abbr. dist) of model selection with emphasis on car \& pedestrian (nuScenes)}
\label{table: adjustment_car_ped}
\end{table}

\begin{figure}[h]
    \includegraphics[scale=0.2]{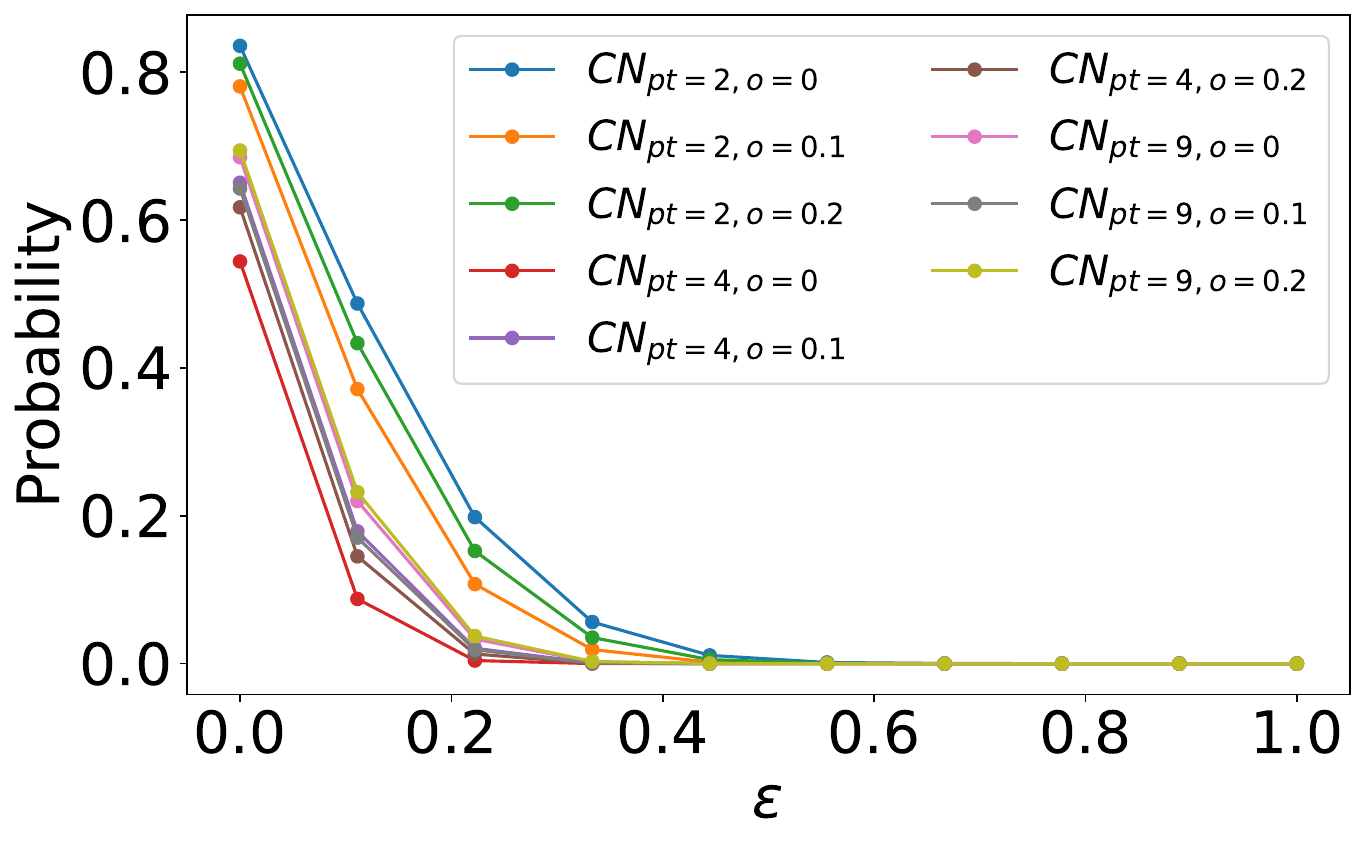}
    \centering
    \caption{The relationship between $\epsilon$ and probability}
    \label{fig: epsilon}
\end{figure}

\textbf{Study of $\epsilon$.} Figure~\ref{fig: epsilon} illustrates how different values of $\epsilon$ affect the computed probabilities. Since the center adjustment relies on these probabilities, choosing an appropriate $\epsilon$ is crucial. An excessively high $\epsilon$ results in probabilities that are too large, diminishing the impact of the adjustment. Conversely, a very low $\epsilon$ causes the adjustment to dominate, reducing the influence of the original distance. From the figure, we observe that setting $\epsilon$ between 0.1 and 0.2 yields the best results for the nuScenes dataset.

Our full experiment for evaluating model selection includes: (1) \textbf{Evaluation of Different Model Combinations.} We conduct experiments using different combinations of models. The results for 9 models and 6 models are presented in Table~\ref{table: adjustment_9} and Table~\ref{table: adjustment_6}, respectively. The corresponding counting results are illustrated in Figure~\ref{fig: adjustment_9} and Figure~\ref{fig: adjustment_6}. \noindent(2) \textbf{Weight adjustment for specific categories.} When generating the center of each model, we assign higher weights to the categories of ``cars" and ``pedestrians". The results of this adjustment are shown in Table~\ref{table: adjustment_car_ped}, with the corresponding counting performance in Figure~\ref{fig: adjustment_9_car_ped}.

With the experiments, we have the following observations: 

\noindent(1) \textbf{Comparison results of using 6 models and 9 models.} Table~\ref{table: adjustment_6} and Figure~\ref{fig: adjustment_6} show the results of using 6 models for model selection. Compared to the results with 9 models, we remove the dominant models from the 9-model set (i.e., $CN_{pt=4}^{o=0}$, $CN_{pt=4}^{o=0.1}$, and $CN_{pt=4}^{o=0.2}$). According to Table~\ref{table: adjustment_6}, after removing the $CN_{pt=4}$ series models, the new dominant model without adjustment becomes $CN_{pt=9}$. However, with the adjustment, the dominant model shifts to $CN_{pt=9}^{o=0.1}$, while $CN_{pt=2}^{o=0.1}$ remains the second dominant model.
From the results in Figure~\ref{fig: adjustment_6}, it is evident that applying adjustments dramatically improves accuracy. When comparing the adjusted results to the best-performing model ($CN_{pt=4}^{o=0.2}$), even the weaker-performing models achieve comparable results through model selection with adjustments. 

\noindent(2) \textbf{Impact of Weight Adjustment on Object Categories.} We also evaluated the effect of adjusting weights for different objects in model selection. Table~\ref{table: adjustment_car_ped} displays the results when the weights for the ``car" and ``pedestrian" categories are doubled, compared to the even-weight approach shown in Table~\ref{table: nus_adjustment}.
From Table~\ref{table: adjustment_car_ped}, the dominant model remains $CN_{pt=4}^{o=0.2}$; however, the second dominant model shifts from $CN_{pt=4}^{o=0}$ to $CN_{pt=9}^{o=0.2}$. As shown in Figure~\ref{fig: adjustment_9_car_ped}, this adjustment significantly improves performance for the ``car" and ``pedestrian" categories, while maintaining stable performance for other object categories.

\subsection{Ablation Studies}
\begin{figure}[t]
    \includegraphics[scale=0.2]{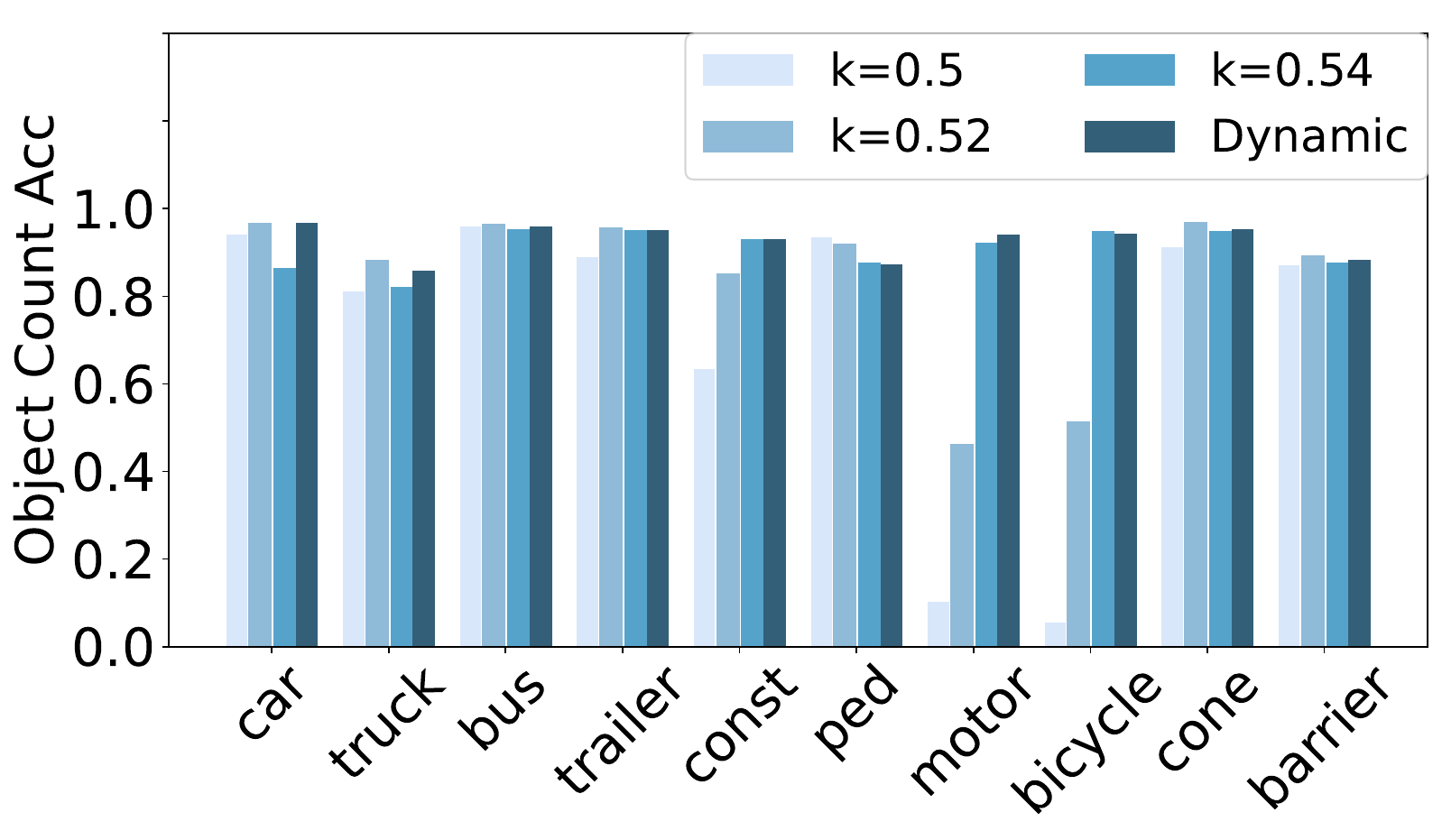}
    \centering
    \caption{Result of Dynamic Thresholding}
    \label{fig: dynamic_threshold}
\end{figure}

\textbf{Dynamic Thresholding.} As explained in Section~\ref{sec: counternet_partition}, we propose using dynamic thresholding to filter out noise for peak detection. We compare the results with and without dynamic thresholding for the model CN-p4o. In our evaluation, we select three fixed thresholds: 0.5, 0.52, and 0.54, for comparison. As shown in Figure~\ref{fig: dynamic_threshold}, using a fixed threshold results in significant deviations across object categories. For instance, the motorcycle and bicycle categories experience extremely low performance with thresholds of 0.5 and 0.52, while the car and truck categories, with a threshold of 0.54, perform poorly. In contrast, dynamic thresholding provide a more balanced performance across all categories, achieving consistently good results.

\noindent\textbf{Weighted loss.}
In Section~\ref{sec: counternet_partition}, we introduce the weighted loss used during model training with partitions. This weighted loss is intended to improve predictions on frames where more objects appear. By referring to the data statistics of the nuScenes dataset in Appendix~\ref{app: nus_distribution}, we observe that the weighted loss contributes more significantly to low-frequency object categories, such as trucks, trailers, and construction vehicles. In contrast, for high-frequency object categories—such as cars, pedestrians, and barriers—the improvement from partitioning is limited.

\begin{figure}[t]
    \includegraphics[scale=0.2]{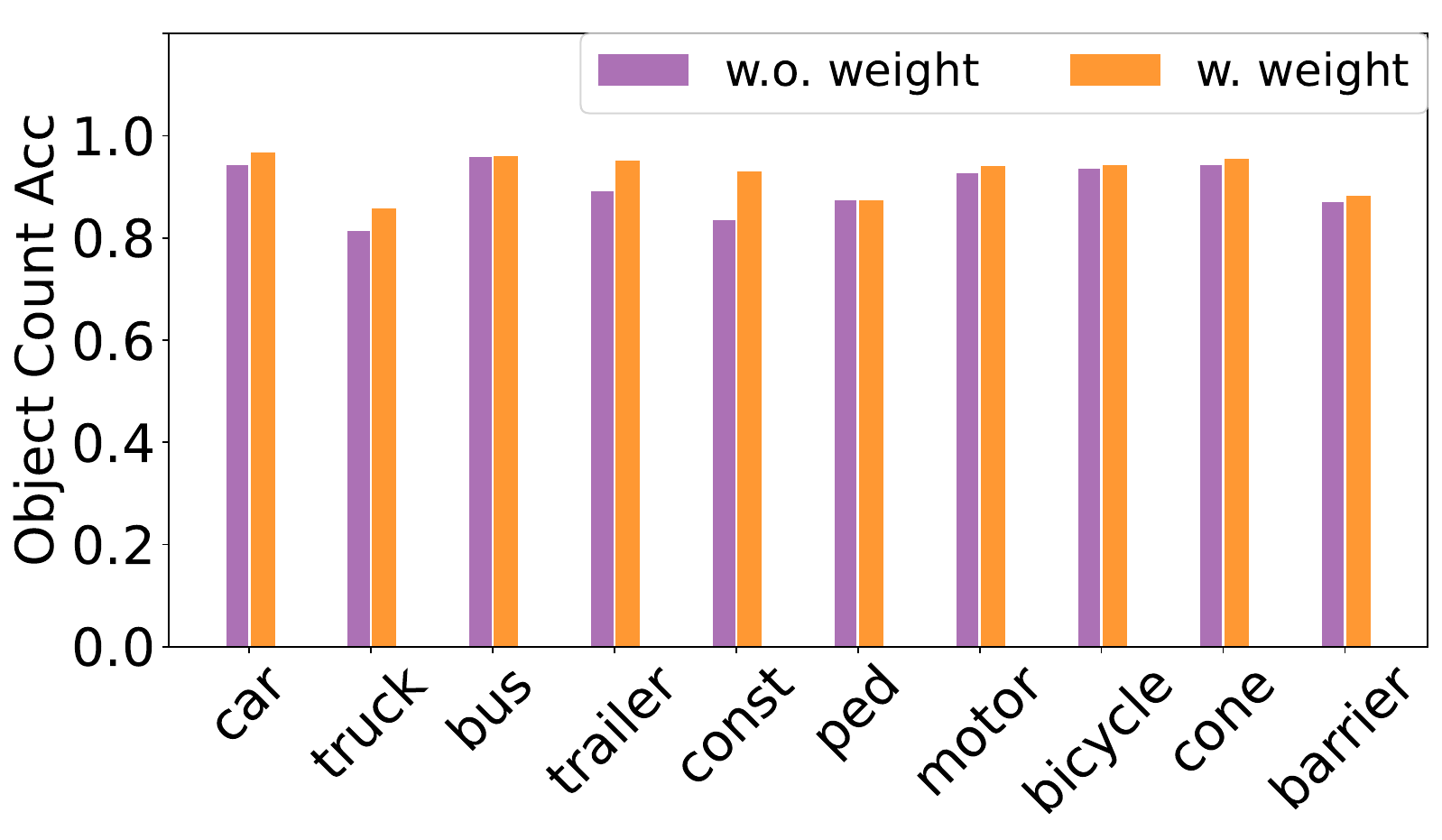}
    \centering
    \caption{Result of Weighted Loss with Partitioning}
    \label{fig: weighted_loss}
\end{figure}

\section{Additional Experiment on KITTI}
\label{app: kitti}

We evaluate our proposed method on the KITTI dataset. We use CenterPoint(CP)~\cite{yin2021centerpoint} as the baseline detection model. 

\subsection{Categorical Query Results}
As shown in Table~\ref{table: kitti_query} and Table~\ref{table: kitti_join}, the performance improvement of our proposed CounterNet ($CN$) on the KITTI dataset is less pronounced compared to the improvements observed on the nuScenes dataset when evaluated against the detection model CenterPoint (CP). For the SELECT-COUNT operation, while the performance gains for the "car" and "cyclist" categories are modest, a noticeable improvement is observed for the "pedestrian" category.

Regarding partitioning and overlap, the results for our proposed method are consistent with those observed on the nuScenes dataset. However, since the scene scale in nuScenes is approximately four times larger than that of KITTI, partitioning has a more negative impact on performance in the KITTI dataset. Nevertheless, the use of overlap partially mitigates this adverse effect.

\begin{figure}[t!]
    \includegraphics[scale=0.2]{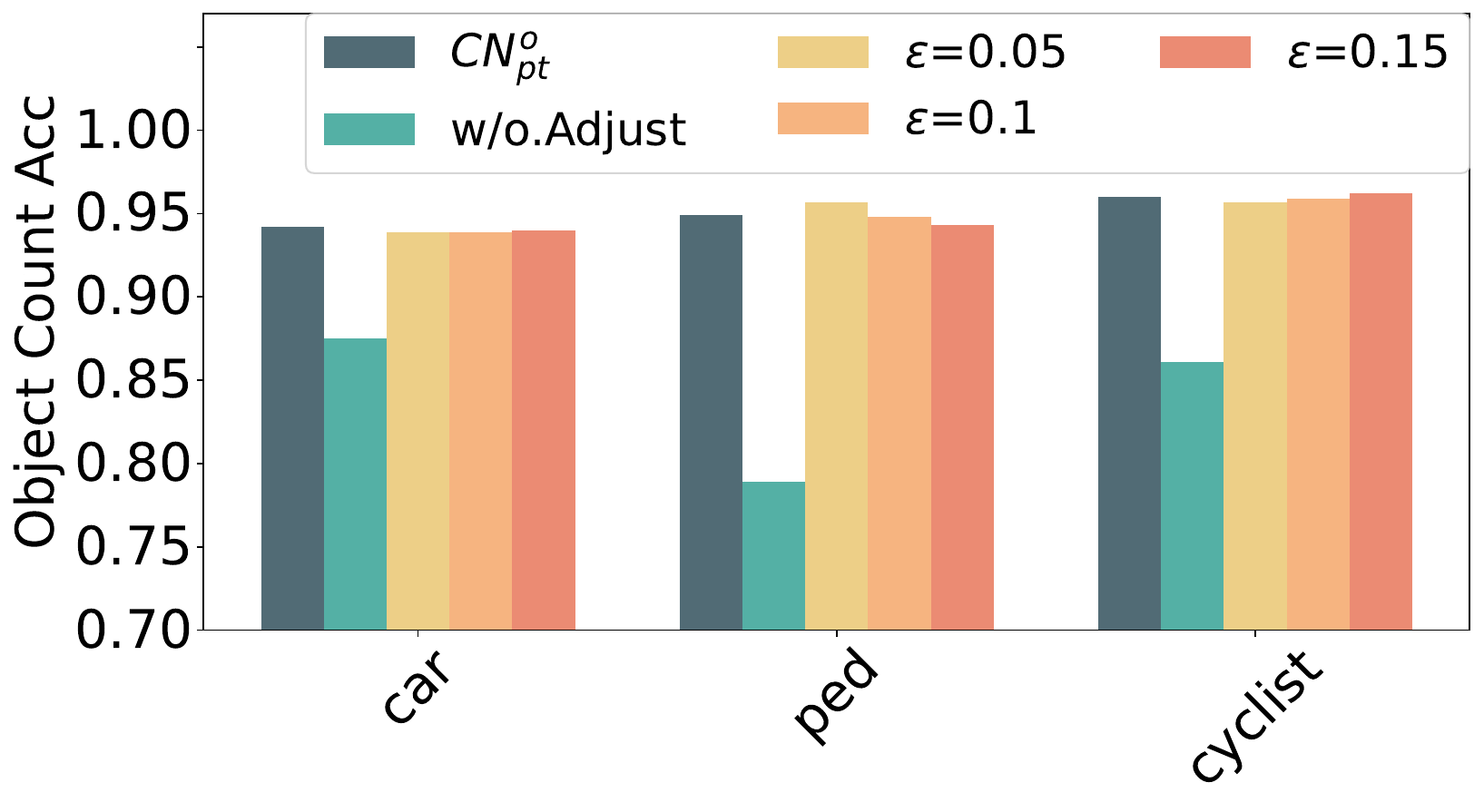}
    \centering
    \caption{Model selection (KITTI)}
    \label{fig: ajustment_kitti}
\end{figure}

\noindent\textbf{Model Selection. }Table~\ref{table: kitti_adjustment} and Figure~\ref{fig: ajustment_kitti} present the results of model selection with six models on the KITTI dataset. Overall, the findings are consistent with those observed on the nuScenes dataset.
As shown in Table~\ref{table: kitti_adjustment}, after applying the adjustment, the frames are primarily concentrated on four partitions with a 0.2 overlap rate ($CN_{pt=4}^{o=0.2}$) and two partitions with a 0.2 overlap rate ($CN_{pt=2}^{o=0.2}$). Furthermore, as illustrated in Figure~\ref{fig: ajustment_kitti}, similar to the nuScenes results, the model selection yields balanced performance across all categories.

\begin{table}[t!]
\centering
\small
\begin{tabular}{|ll|c|cc|cc|cc|}
\hline
\multicolumn{2}{|r|}{Model} & \multicolumn{1}{|r|}{w/o.Adjust} &\multicolumn{2}{c|}{$\epsilon$ = 0.05} & \multicolumn{2}{c|}{$\epsilon$ = 0.1} & \multicolumn{2}{c|}{$\epsilon$ = 0.15} \\ \hline
\multicolumn{1}{|l}{pt}  & \multicolumn{1}{l|}{o}  & dist  & rate                      & dist                    & rate                     & dist                    & rate                       & dist                   \\ \hline
2 &0                         & 937   & 0.956                     & 0                       & 0.838                    & 0                       & 0.672                      & 0                      \\
2&0.2                     & 597   & 0.555                     & 94                      & 0.095                    & 1052                    & 0.005                      & 1648                   \\
4&0                          & 512   & 0.775                     & 0                       & 0.362                    & 0                       & 0.101                      & 61                     \\
4&0.2                     & 165   & 0.307                     & 3675                    & 0.008                    & 2717                    & 0.00002                    & 2060                   \\
9&0                          & 1525  & 0.965                     & 0                       & 0.868                    & 0                       & 0.728                      & 0                      \\
9&0.2                     & 33    & 0.890                     & 0                       & 0.628                    & 0                       & 0.351                      & 0                      \\ \hline
\end{tabular}
\caption{ Data distribution (abbr. dist) of model selection with adjustment (KITTI)}
\label{table: kitti_adjustment}
\end{table}

\section{Results on Waymo dataset}
\label{app: waymo}
We use a subset of the dataset for evaluation, consisting of 18,583 training point clouds and 4,932 for evaluation. The data is sourced from segments prefixed with segment-1 and segment-2.
Table~\ref{table: waymo_query} presents the categorical query results on the Waymo dataset. Overall, the results exhibit similar trends to those observed on the other datasets.

\section{Case studies}
\label{app: case_study}

We evaluate the performance of the actual query across different scenarios and levels of selectivity on the nuScenes dataset. The experimental results assess accuracy, precision, and recall. The key observations are as follows: (1) Accuracy. Our proposed method consistently outperforms the baseline methods in terms of accuracy. However, when precision and recall are considered, the performance varies depending on the specific query. (2) Selectivity. Overall, performance is better in high-selectivity scenarios, whereas low-selectivity cases introduce more errors across all methods.
(3) Query-Specific Analysis. For Query 1 (Q1), CounterNet exhibits a tendency toward higher recall compared to precision, whereas the baselines demonstrate the opposite trend. This indicates that CounterNet achieves higher accuracy within the queried frames but may fail to select some relevant frames. In contrast, the baseline methods tend to select a larger number of frames, capturing more true-positive frames. However, their lower precision indicates reduced accuracy among the selected frames. A similar trend is observed across all other queries.

\end{document}